%% file: main.tex
\documentclass[11pt,a4paper]{article}
\usepackage[hyperref]{acl2021}
\usepackage{times}
\usepackage{latexsym}

\usepackage{microtype}

\aclfinalcopy 

\usepackage{nicefrac}
\usepackage{booktabs}
\usepackage{amsfonts}
\usepackage{bold-extra}
\usepackage{multirow}
\usepackage{multicol}
\usepackage{xspace}
\usepackage{xcolor}

\usepackage{url}
\usepackage{amsmath}
\usepackage{enumitem}
\usepackage{tabularx}
\usepackage{dcolumn,caption}
\usepackage{arydshln}
\usepackage{tikz}
\usepackage{pgfplots}
\pgfplotsset{width=.7\columnwidth}

\usepackage{CJKutf8}

\newcommand{\dotline}{\hdashline[0.5pt/1pt]}

\usepackage{graphicx}
\usepackage{bm}

\newcommand{\abr}[1]{\textsc{#1}}

\newcommand{\our}{\abr{EtA}\xspace}
\newcommand{\ourbert}{\abr{EtA}\,+\,BERT\xspace}

\newcommand{\ourbertplusspiderparser}{\abr{EtA}\,+\,BERT\xspace}
\newcommand{\ourbertpluswikiparser}{\abr{EtA}\,+\,BERT\xspace}
\newcommand{\ourbertplusspiderparserlarge}{\abr{EtA}\,+\,$\text{BERT}_\text{L}$\xspace}

\newcommand{\slsqlg}{SLSQL$_\text{L}$\xspace}
\newcommand{\slsqlgbert}{SLSQL$_\text{L}$\,+\,BERT\xspace}

\newcommand{\slsql}{SLSQL\xspace}
\newcommand{\slsqlbert}{SLSQL\,+\,BERT\xspace}

\newcommand{\baseline}{ALIGN$_\text{P}$\xspace}

\newcommand{\baselinebert}{ALIGN$_\text{P}$\,+\,BERT\xspace}

\newcommand{\alignmodelg}{ALIGN$_\text{L}$\xspace}
\newcommand{\alignmodelgbert}{ALIGN$_\text{L}$\,+\,BERT\xspace}
\newcommand{\alignmodel}{ALIGN\xspace}
\newcommand{\alignmodelbert}{ALIGN\,+\,BERT\xspace}

\newcommand{\ngram}{N-gram Matching\xspace}

\newcommand{\maxpool}{\abr{Contrast}\xspace}
\newcommand{\maxpoolbert}{\abr{Contrast}\,+\,BERT\xspace}

\newcommand{\simmodel}{\abr{Sim}\xspace}
\newcommand{\bertsim}{\abr{Sim}\,+\,BERT\xspace}

\newcommand{\baseparser}{SLSQL$_\text{P}$\xspace}
\newcommand{\baseparserbert}{SLSQL$_\text{P}$\,+\,BERT\xspace}
\newcommand{\baseparserlarge}{SLSQL$_\text{P}$\,+\,$\text{BERT}_\text{L}$\xspace}

\newcommand{\irnet}{IRNet\,+\,BERT\xspace}
\newcommand{\irnetsec}{IRNet v2\,+\,BERT\xspace}

\newcommand{\ratsql}{RATSQL\,+\,$\text{BERT}_\text{L}$\xspace}

\newcommand{\bridgelarge}{BRIDGE\,+\,$\text{BERT}_\text{L}$\xspace}

\newcommand{\vcg}{VCG$^\heartsuit$\xspace}
\newcommand{\elq}{ELQ\,+\,BERT$^\heartsuit$\xspace}
\newcommand{\prior}{Heuristic\xspace}

\newcommand{\wikidata}{\textsc{Squall}\xspace} %
\newcommand{\spiderdata}{\textsc{Spider}-L\xspace}
\newcommand{\wikiorigin}{WTQ\xspace}

\newcommand{\spiderorigin}{Spider\xspace}

\newcommand{\webqspdata}{WebQSP$_\text{EL}$}
\newcommand{\graphquesdata}{GraphQ$_\text{EL}$}

\newcommand{\acclf}{Ex.Match\xspace}
\newcommand{\accexe}{Ex.Acc\xspace}

\newcommand{\colrecall}{$\text{Col}_{R}$\xspace}
\newcommand{\colprec}{$\text{Col}_{P}$\xspace}
\newcommand{\colfscore}{$\text{Col}_{{F}}$\xspace}

\newcommand{\tabrecall}{$\text{Tab}_{{R}}$\xspace}
\newcommand{\tabprec}{$\text{Tab}_{{P}}$\xspace}
\newcommand{\tabfscore}{$\text{Tab}_{{F}}$\xspace}

\newcommand{\entrecall}{$\text{Ent}_{R}$\xspace}
\newcommand{\entprec}{$\text{Ent}_{P}$\xspace}
\newcommand{\entfscore}{$\text{Ent}_{F}$\xspace}

\newcommand{\reftab}[1]{Table~\ref{#1}}
\newcommand{\reffig}[1]{Figure~\ref{#1}}
\newcommand{\refsec}[1]{\S\ref{#1}}

\newcommand*{\affaddr}[1]{#1}
\newcommand*{\affmark}[1][*]{\textsuperscript{#1}}
\newcommand*{\email}[1]{\texttt{\small #1}}

\definecolor{cRed}{RGB}{0,112,255}

\definecolor{cBlue}{RGB}{0,202,150}
\definecolor{cYellow}{RGB}{0,0,0}
\definecolor{cPurple}{RGB}{128,0,128}
\definecolor{cGold}{RGB}{218,165,32}
\newcommand{\red}[1]{\textcolor{cRed}{\textbf{#1}}}
\newcommand{\blue}[1]{\textcolor{cBlue}{\textbf{#1}}}

\newcommand{\redtext}[1]{\textcolor{cRed}{#1}}
\newcommand{\bluetext}[1]{\textcolor{cBlue}{#1}}

\title{Awakening Latent Grounding from Pretrained Language Models\\for Semantic Parsing}

\author{Qian Liu\affmark[\textdagger]{\thanks{~~Work done during an internship at Microsoft Research. The first three authors contributed equally.}}~~, Dejian Yang\affmark[\S], Jiahui Zhang\affmark[\textdagger]$^*$, Jiaqi Guo\affmark[$\lozenge$]$^*$, Bin Zhou\affmark[\textdagger], Jian-Guang Lou\affmark[\S]\\
\affaddr{\affmark[\textdagger]Beihang University, Beijing, China}\\
\affaddr{\affmark[\S]Microsoft Research, Beijing, China}\\
\affaddr{\affmark[$\lozenge$]Xi'an Jiaotong University, Xi'an, China}\\
\affmark[\textdagger]\email{\{qian.liu, 17231043, zhoubin\}@buaa.edu.cn};
\affmark[$\lozenge$]\email{jasperguo2013@stu.xjtu.edu.cn}\\
\affmark[\S]\email{\{dejian.yang, jlou\}@microsoft.com}}

\date{}

\begin{document}
\maketitle
\begin{abstract}
Recent years pretrained language models (PLMs) hit a success on several downstream tasks, showing their power on modeling language.
To better understand and leverage what PLMs have learned, several techniques have emerged to explore syntactic structures entailed by PLMs.
However, few efforts have been made to explore grounding capabilities of PLMs, which are also essential.
In this paper, we highlight the ability of PLMs to discover which token should be grounded to which concept, if combined with our proposed erasing-then-awakening approach.
Empirical studies on four datasets demonstrate that our approach can awaken latent grounding which is understandable to human experts, even if it is not exposed to such labels during training.
More importantly, our approach shows great potential to benefit downstream semantic parsing models.
Taking text-to-SQL as a case study, we successfully couple our approach with two off-the-shelf parsers, obtaining an absolute improvement of up to $9.8$\%.

\end{abstract}

\section{Introduction}

Recent breakthroughs of \textbf{P}retrained \textbf{L}anguage \textbf{M}odels (\textbf{PLM}s) such as BERT \citep{devlin-etal-2019-bert} and GPT3 \citep{brown2020language} have demonstrated the effectiveness of self-supervised learning for a range of downstream tasks.
Without being guided by structural information in training, PLMs show the potential for learning implicit syntactic structures and language semantic, which can be transferred to other tasks.
To better understand and leverage what PLMs have learned, several work has emerged to probe or induce syntactic structures from PLMs.
According to prior studies \citep{anna-etal-2021-primer}, most existing work focuses on syntactic structures such as part of speech \citep{liu-etal-2019-linguistic}, constituency tree \citep{wu-etal-2020-perturbed} and dependency tree \citep{hewitt-manning-2019-structural,jawahar-etal-2019-bert}, paying much less attention on language semantics \citep{tenney2018you}.
However, as well known, semantic information is essential for high-level tasks like machine reading comprehension \citep{wang-jiang-2019-explicit}.

\begin{figure}[t]
        \centering
        \includegraphics[width=1.0\linewidth]{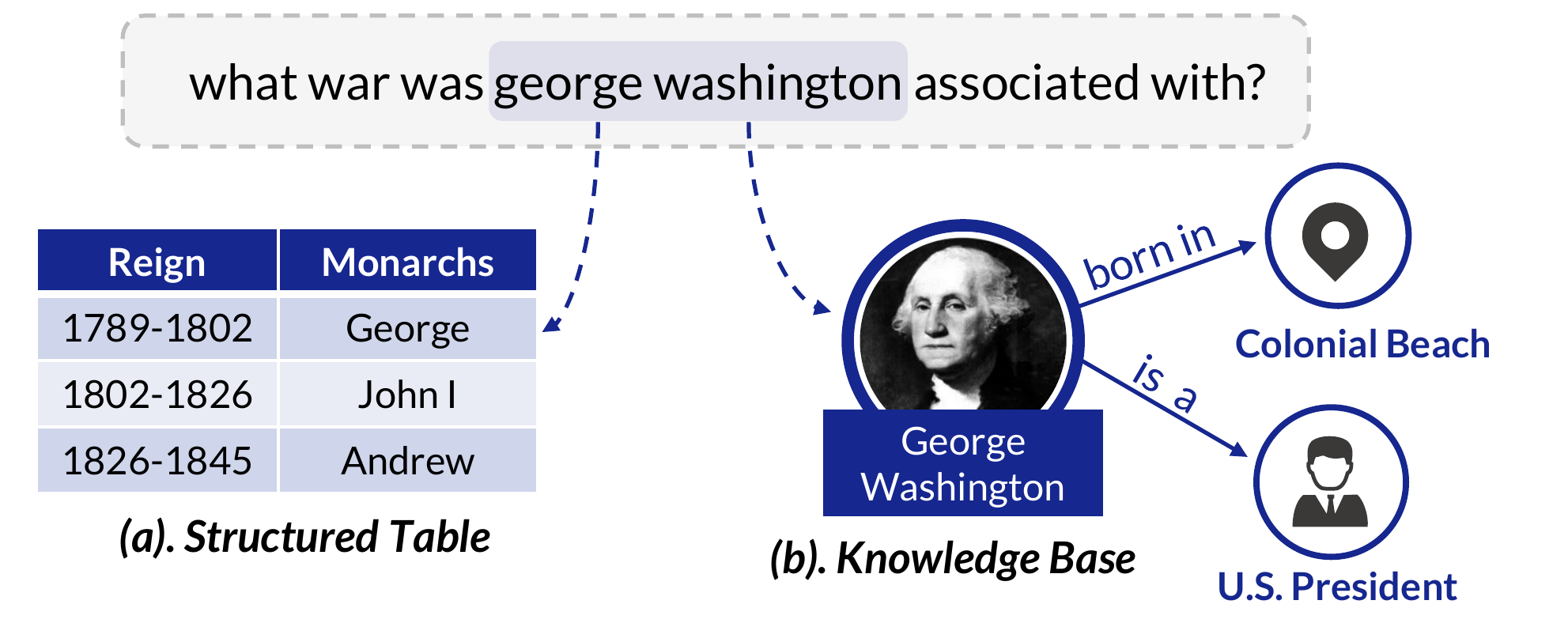}
        \caption{Typical scenarios for grounding, here the linguistic tokens ``george washington'' can be grounded into different real-world concepts.}
        \label{fig:intro_grounding}
\end{figure}

\begin{figure*}[t]
        \centering
        \includegraphics[width = 0.8\linewidth]{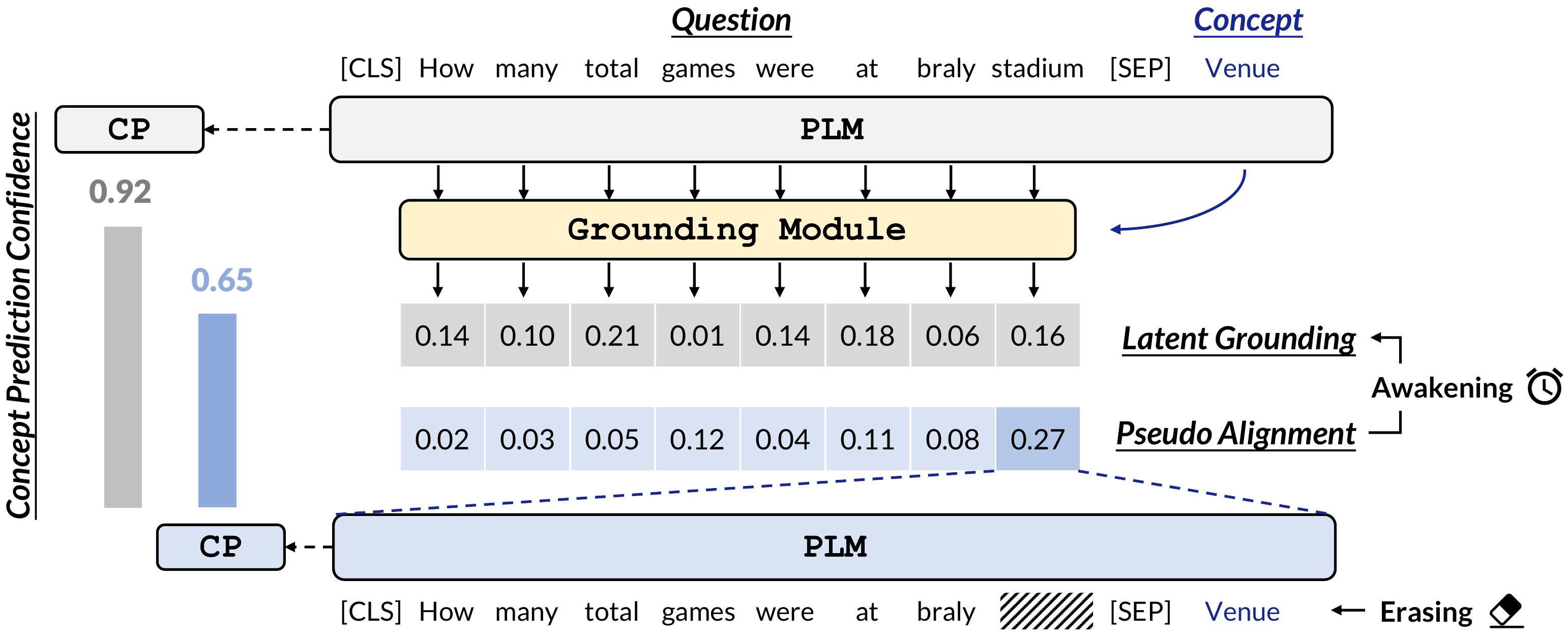}
        \caption{The illustration of \our, which consists of a PLM module, a \textbf{C}oncept \textbf{P}rediction (CP) module and a grounding module.
        Two models (gray and blue) are drawn here for illustration purposes, and they are indeed the same.
        The model training involves three steps:
        (1) The concept prediction module is trained to predict the confidence of any concept occurring in a given question (\textbf{Left}).
        (2) The erasing mechanism erases tokens in the question sequentially, feeds them into CP, and obtains the confidence differences (e.g., $0.92-0.65=0.27$) as the pseudo alignment.
        Here we only demonstrate the process related to ``stadium'' (\textbf{Bottom Right}).
        (3) The pseudo alignment is employed to awaken the latent grounding, i.e., to supervise the grounding module (\textbf{Top Right}).
        We show only one concept ``Venue'' for the sake of brevity, which in practice is a sequence of concepts.
        }
        \label{fig:model}
\end{figure*}

Regarding to language semantics, an important branch is grounding, which is overlooked by most previous work.
Broadly speaking, grounding means ``\textit{connecting linguistic symbols to real-world perception or actions}'' \citep{roy2005grounding}.
It is generally thought to be important for a variety of tasks, such as video descriptions \citep{DBLP:conf/cvpr/ZhouKCCR19}, visual question answering \citep{DBLP:conf/cvpr/ZhuGBF16} and semantic parsing \citep{guo-etal-2019-towards}.
In this paper, we focus on single-modal scenarios, where grounding refers more specifically to mapping linguistic tokens into a real-world concept described in natural language.
As shown in \reffig{fig:intro_grounding}, ``george washington'' can be grounded into either a cell value in a structured table, or an entity in knowledge bases.

In single-modal scenarios, grounding is especially important for semantic parsing, the task of translating a natural language sentence into its corresponding executable logic form.
For earlier work, grounding is essential since earlier work almost conceptualized semantic parsing as grounding an utterance to a task-specific meaning representation \citep{zelle1996learning,zettlemoyer2005learning,liang-etal-2013-learning,cheng-etal-2017-learning}.
As for modern approaches based on the encoder-decoder architecture, grounding also plays an important role and considerable work has demonstrated the positive effect of it \citep{guo-etal-2019-towards,dong-etal-2019-data,qian2020how,wang-etal-2020-relational,chen-etal-2020-tale}.
Despite its success, existing grounding methods mainly relied on heavy manual efforts like high-quality lexicons \citep{reddy-etal-2016-transforming} or ad-hoc heuristic rules like n-gram matching \citep{guo-etal-2019-towards}, suffering from poor flexibility.
To explore more flexible methods, researchers recently tried a data-driven way: they collected grounding annotations as supervision to train grounding models \citep{li-etal-2020-efficient,lei-etal-2020-examining,shi-etal-2020-potential}.
However, this modeling flexibility in their approaches requires expensive annotations of grounding, which most of the time are not available.

To alleviate the above issues, we present a novel approach \textbf{E}rasing-\textbf{t}hen-\textbf{A}wakening (\our)\footnote{Our code is available at \url{https://github.com/microsoft/ContextualSP}}.
It is inspired by recent advances in interpretable machine learning \citep{samek2016evaluating}, where the importance of individual pixels can be quantified with respect to the classification decision.
Similarly, our approach firstly quantifies the contribution of each word with respect to each concept, by erasing it and probing the variation of concept prediction decisions (elaborated later).
Then it employs these contributions as pseudo labels to awaken latent grounding from PLMs.
In contrast to prior work, our approach only needs supervision of concept prediction, which can be easily derived by downstream tasks (e.g., text-to-SQL) instead of full grounding supervision.
Empirical studies on four datasets demonstrate that our approach can awaken latent grounding which is understandable to human experts.
It is highly non-trivial because our approach is not exposed to any human-annotated grounding label in training.
More importantly, we find that the grounding can be easily coupled with downstream models to boost their performance, and the absolute improvement is up to $9.8\%$.
In summarization, our contribution is as three-fold:
\begin{enumerate}
    \item To the best of our knowledge, we are the first one to highlight and demonstrate the possibility of awakening latent grounding from PLMs.
    \item We propose a novel weakly supervised approach erasing-then-awakening, to awaken latent grounding from PLMs. Empirical studies on four datasets demonstrate that our approach can awaken latent grounding which is understandable to human experts.
    \item Taking text-to-SQL as a case study, we successfully couple our approach with two off-the-shelf parsers. Experimental results on two benchmarks show the effectiveness of our approach on boosting downstream performance.
\end{enumerate}

\section{Method: Erasing-then-Awakening}\label{sec:method}

In the task of grounding, we are given a question $\mathbf{x} = \langle x_1,\cdots,x_N \rangle$ and a concept set $\mathcal{C}=\{c_1,\cdots,c_K\}$, where each concept consists of several tokens.
The goal of grounding is to find out tokens (also known as mentions) in $\mathbf{x}$ which are relevant to concepts in $\mathcal{C}$.
Generally, the grounding procedure learns to create a $N{\times}K$ matrix, which we call \textit{latent grounding}.
In some cases, a set of pairs is needed, of which each one explicitly shows a token and a concept is grounded.
We call this kind of pairs as \textit{grounding pairs} below.

As illustrated in Figure~\ref{fig:model}, our model consists of a PLM module, a CP module and a grounding module.
In this section, we first present the training procedure of \our, which at a high-level involves three steps:
(1) Train an auxiliary concept prediction module.
(2) Erase tokens in a question to obtain the concept prediction confidence differences as pseudo alignment.
(3) Awaken latent grounding from PLMs by applying pseudo alignment as supervision.
Then we introduce the procedure to produce grounding pairs in inference.

\subsection{Training a Concept Prediction Module} \label{sec:concept}

Given $\mathbf{x}$ and $\mathcal{C}$, the goal of the concept prediction module is to identify if each concept $c_k\in \mathcal{C}$ is mentioned or not in the question $\mathbf{x}$.
Although it does not seem to be directly related to grounding, it is a pre-requisite for the erasing mechanism, which will be elaborated later.
As for $c_k$'s supervision $l_k\in\{0,1\}$, it is the weak supervision \our relies on, and can be readily obtained through downstream task signals.
Taking text-to-SQL as an illustration, each database schema (i.e., table, column and cell value) in an annotated SQL can be considered as mentioned in the question ($l_k=1$), with others as negative examples ($l_k=0$).

Once the supervision is prepared, the CP module is trained to conduct binary classification over the representation of each concept.
As done in previous work \citep{Hwang2019ACE}, we first concatenate the question and all concepts into a sequence as input to the PLM module.
As illustrated in Figure~\ref{fig:model}, the input sequence starts with \texttt{\small[CLS]}, with the question and each concept being separated by \texttt{\small[SEP]}.
Then, the sequence is fed into the PLM module to produce deep contextual representations over each position.
Denoting $\langle \mathbf{q}_1,\mathbf{q}_2,...,\mathbf{q}_N \rangle$ and $\langle \mathbf{e}_1,\mathbf{e}_2,...,\mathbf{e}_K \rangle$ as the token representations and concept representations, they can be obtained by:
\begin{equation}
    \small
     \{\!\mathbf{q}_n\}_{n\!=\!1}^{N}, \{\!\mathbf{e}_k\}_{k\!=\!1}^{K}\!=\!\text{PLM}\big(\texttt{[CLS]}\!,\!\mathbf{x},\!\{\!\texttt{[SEP]},c_k\}_{k\!=\!1}^{K}\big),
\end{equation}
where $\mathbf{q}_n$ and $\mathbf{e}_k$ correspond to the representations at the position of $n$-th question token and the first token in $c_k$ respectively.
Finally, each concept representation $\mathbf{e}_k$ is passed to a classifier to predict if it is mentioned in $\mathbf{x}$ as:
\begin{equation}
    p_k = \mathtt{Sigmoid}(\mathbf{W}_l\,\mathbf{e}_k),
\end{equation}
where $\mathbf{W}_l$ is a learnable parameter.
$p_k$ is the probability of $c_k$ mentioned in the question, which is referred to by \textit{concept prediction confidence} below.

\subsection{Erasing Question Tokens} \label{sec:erasing}

Once the concept prediction module is converged, we apply an erasing mechanism to assist in the following awakening phase.
It follows a similar idea from the interpretable document classification \citep{DBLP:journals/corr/ArrasHMMS16a}, where a word is considered important for the document classification if removing it and classifying the modified document results in a strong decrease of the classification score.
In our case, a token is considered highly relevant to certain concepts if there is a large drop in these concept prediction confidences after erasing the token.
Therefore, we need the above mentioned concept prediction module to provide a reasonable concept prediction confidence.

Concretely, as shown in Figure~\ref{fig:model}, the erasing mechanism erases the input sequentially, and feeds each erased input into the PLM module and the subsequent CP module.
For example, with $x_n$ being substituted by a special token \texttt{\small [UNK]}, we can obtain an erased input as \texttt{\small [CLS]}$,x_1,\cdots,x_{n-1}$,\texttt{\small [UNK]}$,x_{n+1},\cdots,c_K$.
Denoting $\hat{p}_{n,k}$ the concept prediction confidence for $c_k$ after erasing $x_n$, we believe the difference between $\hat{p}_{n,k}$ and $p_k$ reveals $c_k$'s relevance to $x_n$ from a PLM's view.
The confidence difference $\Delta_{n,k}$ can be obtained by ${\Delta}_{n,k}=l_k{\cdot}\max(0, p_k-\hat{p}_{n,k})$.
Repeating the above procedure on the input question sequentially, $\Delta\,{\in}\,\mathbb{R}^{N{\times}K}$ is filled completely.

\subsection{Awakening Latent Grounding}\label{sec:awakening_method}

As mentioned above, we believe $\Delta$ reflects the relevance between each token and each concept from a PLM's view.
Therefore, we could directly use $\Delta$ as \our's output.
However, according to our preliminary study, the method performs poorly and cannot produce high-quality alignment\footnote{More experimental results can be found in \refsec{sec:model_analysis}.}.
Different from directly using $\Delta$, we employ it to ``awaken'' the latent grounding.
To be specific, we introduce a grounding module upon representations of the PLM module and train it using $\Delta$ as pseudo labels (i.e., pseudo alignment).
The grounding module first obtains grounding scores $g_{n,k}$ between each question token $x_n$ and each concept $c_k$ based on their deep contextual representations $\mathbf{q}_n$ and $\mathbf{e}_k$ as:
\begin{equation}
    g_{n,k} = \frac{\mathbf{W}_e\mathbf{e}_{k}\cdot({\mathbf{W}_q\mathbf{q}_n})^T}{\sqrt{d}},
\end{equation}
where $\mathbf{W}_e,\mathbf{W}_q$ are learnable parameters and $d$ is the dimension of $\mathbf{e}_k$.
Then it normalizes the grounding scores into latent grounding $\boldsymbol{\alpha}$ as:
\begin{equation}
    \boldsymbol{\alpha}_{n,k} = \frac{\exp(g_{n,k})}{\sum\nolimits_{i}\exp(g_{i,k})}.
\end{equation}
Finally, the grounding module is trained to maximize the likelihood with ${\Delta}$ as the weight:
\begin{equation}
\sum\limits_{n} \sum\limits_{k} {\Delta}_{n,k}\cdot\log{{\boldsymbol{\alpha}}_{n,k}}.
\end{equation}

\subsection{Producing Grounding Pair}\label{method:producing}

Repeating erasing and awakening iteratively for epochs until the grounding module converges, we can readily produce grounding pairs.
Formally, we aim to obtain a set of pairs, where each pair $\langle x_n, c_k\rangle$ indicates that $x_n$ is grounded to $c_k$.
Noticing $c_k$ may contain several tokens, we keep all probabilities in $\boldsymbol{\alpha}_{\cdot,k}$ which exceeds $\tau / |c_k|$, where $\tau$ is a threshold and $|c_k|$ is the number of tokens in $c_k$.
Also, taking into account that $x_n$ should be grounded to only one concept, we keep only the highest probability over $\boldsymbol{\alpha}_{n,\cdot}$.
Finally, for each pair $\langle x_n,c_k \rangle$, it is thought to be a grounding pair if $\boldsymbol{\alpha}_{n,k}$ is kept and $p_k \geq 0.5$, otherwise it is not.

\section{Experiments}

\begin{table*}[t]
  \centering
  \small
  \begin{tabular}{lcccccccccc}
    \toprule
      \multirow{2}{*}{Model} & \multicolumn{6}{c@{\hspace{3.5pt}}}{\spiderdata} & \multicolumn{3}{c}{\wikidata} \\
      \cmidrule(lr){2-7} \cmidrule(lr){8-10}
     & \colprec & \colrecall & \colfscore & \tabprec & \tabrecall & \tabfscore & \colprec & \colrecall & \colfscore  \\
    \midrule
    \ngram & $61.4$ & $69.1$ & $65.1$ & $78.2$ & $69.6$ & $73.6$ & $71.6$ & $50.8$ & $59.4$\\
    \bertsim & $16.6$ & $8.0$ & $10.8$ & $8.5$ & $11.6$ & $9.8$ & $13.9$ & $18.0$ & $15.7$ \\ 
    \maxpoolbert & $83.7$ & $68.4$ & $75.3$ & $\mathbf{84.0}$ & $76.9$ & $80.3$ & $47.9$ & $31.2$ & $37.8$ \\
    \ourbert & $\mathbf{86.1}$ & $\mathbf{79.3}$ & $\mathbf{82.5}$ & $81.1$ & $\mathbf{85.3}$ & $\mathbf{83.1}$ & $\mathbf{77.3}$ & $\mathbf{62.4}$ & $\mathbf{69.0}$ \\
    \midrule
    \slsqlgbert\!$^\heartsuit$ \citep{lei-etal-2020-examining} & $82.6$ & $82.0$ & $82.3$ & $80.6$ & $84.0$ & $82.2$ & -- & -- & -- \\
    \alignmodelgbert\!$^\heartsuit$ \citep{shi-etal-2020-potential} & -- & -- & -- & -- & -- & -- & $79.2$ & $72.8$ & $75.8$ \\
    \bottomrule
  \end{tabular}
    \caption{Experimental results on schema linking dev sets.
    $^\heartsuit$ means the model uses schema linking supervision, while other learnable models use weak supervision.
    +BERT means using BERT as encoder, the same for \reftab{tb:webqsp-linking-res}.
    }
  \label{tb:spider-linking-res}
\end{table*}

In this section, we conduct experiments to evaluate if the latent grounding awakened by \our is understandable to human experts.
Here we accomplish the evaluation by comparing the grounding pairs produced by \our with human annotations.

\subsection{Experimental Setup} \label{section:interpretable_ex_setup}

\begin{table*}[t]
  \centering
  \small
  \begin{tabular}{lcccccc}
      \toprule
      \multirow{2}{*}{Model} & \multicolumn{3}{c@{\hspace{3.5pt}}}{\webqspdata} & \multicolumn{3}{c}{\graphquesdata (zero-shot)} \\
      \cmidrule(lr){2-4} \cmidrule(lr){5-7}
      & \entprec & \entrecall & \entfscore & \entprec & \entrecall & \entfscore \\
    \midrule
    \prior \citep{sorokin-gurevych-2018-mixing} & $30.2$ & $60.8$ & $40.4$ & - & - & - \\
    \ourbert & $\mathbf{76.6}$ & $\mathbf{72.5}$ & $\mathbf{74.5}$ & $\mathbf{43.1}$ & $\mathbf{42.1}$ & $\mathbf{42.7}$ \\
    \midrule
    \vcg \citep{sorokin-gurevych-2018-mixing} & $82.4$ & $68.3$ & $74.7$ & $54.1$ & $30.6$ & $39.0$ \\
    \elq \citep{li-etal-2020-efficient} & $90.0$ & $85.0$ & $87.4$ & $60.1$ & $57.2$ & $58.6$ \\
    \bottomrule
  \end{tabular}
    \caption{Experimental results on entity linking test sets.
    $^\heartsuit$ means the model uses entity linking supervision from \webqspdata, while \our uses the weak supervision derived from WebQSP.
    Following previous work \citep{sorokin-gurevych-2018-mixing}, we use \graphquesdata only in the evaluation phase to test the generalization ability of our model.}
  \label{tb:webqsp-linking-res}
\end{table*}

\paragraph{Datasets} We select two representative grounding tasks where human annotations are available: \textit{schema linking} and \textit{entity linking}.
Schema linking is to ground questions into database schemas, while entity linking is to ground questions into entities of knowledge bases.
For schema linking, we select \spiderdata \citep{lei-etal-2020-examining} and \wikidata \citep{shi-etal-2020-potential} as our evaluation benchmarks.
As mentioned in \refsec{sec:concept}, the supervision for our model is obtained from SQL queries.
As for entity linking, we select \webqspdata and \graphquesdata \citep{sorokin-gurevych-2018-mixing}.
The supervision for our model is obtained from SPARQL queries in a similar way.

\paragraph{Evaluation}

For schema linking, as done in previous work \citep{lei-etal-2020-examining}, we report the micro-average precision, recall and F1-score for both columns (\colprec, \colrecall, \colfscore) and tables (\tabprec, \tabrecall, \tabfscore).
For entity linking, we report the weak matching precision, recall and F1-score for entities (\entprec, \entrecall, \entfscore).
The weak matching metric is a commonly used metric in previous work \citep{sorokin-gurevych-2018-mixing}, which considers a prediction as correct whenever the correct entity is identified and the predicted mention boundary overlaps with the ground truth boundary.
More details can be seen in \refsec{appendix:evaluation}.

\paragraph{Baselines} 
For schema linking, we consider four strong baselines.
(1) \textbf{\ngram} enumerates all n-gram ($n \leq 5$) phrases in a natural language question, and links them to database schemas by fuzzy string matching.
(2) \textbf{\simmodel} computes the dot product similarity between each question token and schema using their PLM representations without fine-tuning, to explore grounding capacities of unawakened PLMs.
(3) \textbf{\maxpool} learns by comparing the aggregated grounding scores of mentioned schemas with unmentioned ones in a contrastive learning style, as done in \citet{liu-etal-2020-impress}. Concretely, in training, \maxpool is trained to accomplish the same concept prediction task as our approach. With a similar architecture to the Receiver used in \citet{liu-etal-2020-impress}, it first computes the similarity score between each token and each concept, and then uses max pooling to aggregate the similarity scores of a concept over an utterance into a concept prediction score. Finally, a margin-based loss is used to encourage the baseline to give higher concept prediction scores on mentioned concepts than unmentioned concepts.
(4) \textbf{\slsqlg \& \alignmodelg}.
\slsqlg(\alignmodelg) is a learnable schema linking module\footnote{\slsql and \alignmodel use multi-task learning to simultaneously learn schema linking and SQL generation.} proposed in \slsql(\alignmodel).
Unlike our method, these two methods are trained with the full schema linking supervision.
Please refer to \citet{shi-etal-2020-potential} and \citet{lei-etal-2020-examining} for more details.
Notably, for baselines which require a threshold, we tuned their thresholds based on dev sets for fair comparison.

For entity linking, we compare \our with three powerful methods.
(1) \textbf{\prior} picks the most frequent entity among the candidates found by string matching over Wikidata.
(2) \textbf{VCG}~\citep{sorokin-gurevych-2018-mixing} aggregates and mixes contexts of different granularities to perform entity linking.
(3) \textbf{ELQ}~\citep{li-etal-2020-efficient} uses a bi-encoder to perform entity linking in one pass, achieving state-of-the-art performance on \webqspdata and \graphquesdata.
VCG and ELQ utilize entity linking supervision in training, while \our does not.

\paragraph{Implementation}

For schema linking we follow the procedure in \refsec{method:producing} to produce grounding pairs to evaluate, while for entity linking we further merge adjacent grounding pairs to produce span-level grounding pairs.
We implement \our in Pytorch \citep{NEURIPS2019_bdbca288}.
With respect to PLMs in experiments, we use the uncased BERT-base (BERT)\footnote{Our approach is theoretically applicable to different PLMs. In this paper, we chose BERT as a representative and we leave exploration of different PLMs for future work.} and BERT-large (BERT$_\text{L}$) from Transformers library \citep{wolf-etal-2020-transformers}.
As for the optimizer, we employ AdamW \citep{loshchilov2018decoupled}.
More details (e.g., learning rate) of each experiment can be found in \refsec{appendix:implement_grounding}.

\subsection{Experimental Results}

\begin{table*}
    \small
    \centering
    \begin{tabularx}{\textwidth}{ll}
    \toprule
        \textbf{Error Type} & \multicolumn{1}{c}{\textbf{Example Error}} \\
    \midrule
        \textbf{Missed Grounding (43.1\%)} & How many \blue{points} did arnaud demare receive? \\
        & \verb|GOLD:| \blue{points}$\to$ ``\verb|UCI world tour points|"\;\;\;\;\;\;\verb|PRED:| \\
    \midrule
        \textbf{Technically Correct (21.0\%)} & Total \blue{population} of millbrook first \red{nation}? \\
        & \verb|GOLD:| \blue{population}$\to$ ``\verb|Population|" \\
        & \verb|PRED:| \blue{population}$\to$ ``\verb|Population|"; \red{nation}$\to$ ``\verb|Community|" \\
    \midrule
        \textbf{Partially Correct (15.8\%)} & Who was \red{the first} \blue{winning captain}? \\
        & \verb|GOLD:| \red{the first}$\to$ ``\verb|Year|"; \blue{winning captain}$\to$ ``\verb|Winning Captain|"\\
        & \verb|PRED:| \red{first}$\to$ ``\verb|Year|"; \blue{winning captain}$\to$ ``\verb|Winning Captain|" \\
    \midrule
       \textbf{Wrong Grounding (10.1\%)} & Were the \red{matinee} and evening performances held \blue{earlier} than the 8th anniversary? \\
        & \verb|GOLD:| \blue{earlier}$\to$ ``\verb|Date|" \\
        & \verb|PRED:| \red{matinee}$\to$ ``\verb|Performance|"; \blue{earlier}$\to$ ``\verb|Date|" \\
    \bottomrule
    \end{tabularx}
    \caption{Four main error types made by \our along with their proportions on \wikidata dataset.}
    \label{tab:error_analysis}
\end{table*}

Table~\ref{tb:spider-linking-res} shows the experimental results on the schema linking task.
As shown, our method outperforms all weakly supervised methods and heuristic-based methods by a large margin.
For example, on \spiderdata, \ourbert achieves an absolute improvement of $7.2$\% \colfscore and $2.8$\% \tabfscore over the best baseline \maxpool.
The same conclusion can be drawn from the experimental results on the entity linking task shown in Table~\ref{tb:webqsp-linking-res}.
For instance, \ourbert can obtain a high \entfscore up to $74.5$\% on \webqspdata, which is a satisfying performance for downstream tasks.
All results above demonstrate the superiority of our approach on awakening latent grounding from PLMs.
With respect to the reason that PLMs work well on both schema linking and entity linking, it may be because both schema linking and entity linking require text-based semantic matching (e.g., synonyms), which PLMs excel at.

Furthermore, it is very surprising that although not trained under fine-grained grounding supervision, our model is comparable with or slightly worse than the fully supervised models across datasets.
For instance, on \spiderdata, our model exceeds the fully supervised baseline \slsqlg by $0.9$ points on \tabfscore.
On \wikidata, our model holds a slightly worse performance than the fully supervised baseline \alignmodelg.
It is highly nontrivial since \maxpool, the best weakly supervised baseline on \spiderdata, is far from the fully supervised model on \wikidata, while our model has only a small drop.
Besides, on \webqspdata and \graphquesdata, although our model is inferior to the state-of-the-art model ELQ, it also achieves a comparable performance with the fully supervised baseline VCG.
These results provide strong evidence that PLMs do have very good grounding capabilities, and our approach can awaken them from PLMs.

\subsection{Model Analysis}\label{sec:model_analysis}

In this section, we try to answer four interesting research questions via a thorough analysis: \textbf{RQ1.} Does the grounding capability come mainly from the PLM? \textbf{RQ2.} Is the awakening phase necessary? \textbf{RQ3.} Do larger PLMs have better grounding capabilities? \textbf{RQ4.} What are the remaining errors?

\paragraph{RQ1}
There is a long term debate in literature about if knowledge is primarily learned by PLMs, when extra parameters are employed in analysis \citep{hewitt-liang-2019-designing}.
Similarly, since our approach depends on extra modules (e.g., grounding module), it faces the same dilemma: how can we know whether the latent grounding is learnt from PLMs or extra modules?
Therefore, we apply our approach to a randomly initialized Transformer encoder \citep{DBLP:conf/nips/VaswaniSPUJGKP17}, to probe the grounding capability of a model that has not been pretrained.
To make it comparable, the encoder has the same architecture as BERT.
However, it only gets a $40$\% \colfscore on \wikidata, not even as good as the N-gram baseline.
Considering it contains the same extra modules as \ourbert, the huge gap between it and \ourbert supports the opinion that the latent grounding is mainly learnt from PLMs.
Meanwhile, one concern shared by our reviewers is the risk of supervision exposure during training of the concept prediction module.
In other words, our approach may ``steal'' some supervision in the concept prediction module to achieve good performance on grounding.
However, the above experiment demonstrates that a non-pretrained model is far from strong grounding capability even with the same concept prediction module.
We hope the finding will alleviate the concern.

\input{tk_plot}

\paragraph{RQ2}
As mentioned in \refsec{sec:awakening_method}, the pseudo alignment $\Delta$ can also be employed as the model prediction.
Therefore, we conduct experiments to verify if our proposed awakening phase is necessary.
As shown in \reffig{fig:self_teaching}, even with various normalization methods (e.g., $\mathtt{Softmax}$), $\Delta$ does not produce satisfactory alignment.
In contrast, our model consistently performs well.
To investigate deeper, we conduct a careful analysis on $\Delta$, and we are surprised to find that values of $\Delta$ are generally small and not as significantly different with each other as we would expect.
Therefore, we believe the success of our approach stems from the fact that it encourages the grounding module to capture subtle differences and strength them.

\paragraph{RQ3}

We apply our approach on BERT-large ($\text{BERT}_\text{L}$) and conduct experiments on \spiderdata.
The results show $\text{BERT}_\text{L}$ brings an improvement of $2.5$\% \colfscore and $0.5$\% \tabfscore, suggesting the possibility of awakening better latent grounding from larger PLMs.
Nevertheless, the improvement may also come from more parameters, so the conclusion needs further investigation.

\paragraph{RQ4}\label{section_error_analysis}

We manually examine $20$\% of our model's errors on the \wikidata dataset and summarize four main error types: (1) missed grounding - where our model did not ground any token to a concept, (2) technically correct - where our model was technically correct but the annotation was missing, (3) partially correct - where our model did not find all tokens of a concept, (4) wrong grounding - where the model produced incorrect grounding.
As shown in \reftab{tab:error_analysis}, only a small fraction of errors are wrong grounding, indicating that the main challenge of our approach is recall rather than precision.

\section{Case Study: Text-to-SQL}\label{sec:case_study}

The \our model is proposed for general-purpose uses and intends to enhance different downstream semantic parsing models.
To verify it, we take the text-to-SQL task as a case study.
In this section, we first present a general solution to couple \our with different text-to-SQL parsers.
Then, we conduct experiments on two off-the-shelf parsers to verify the effectiveness of \our.

\subsection{Coupling with Text-to-SQL Parsers}

\begin{figure}[t]
    \centering
    \includegraphics[width=1.0\columnwidth]{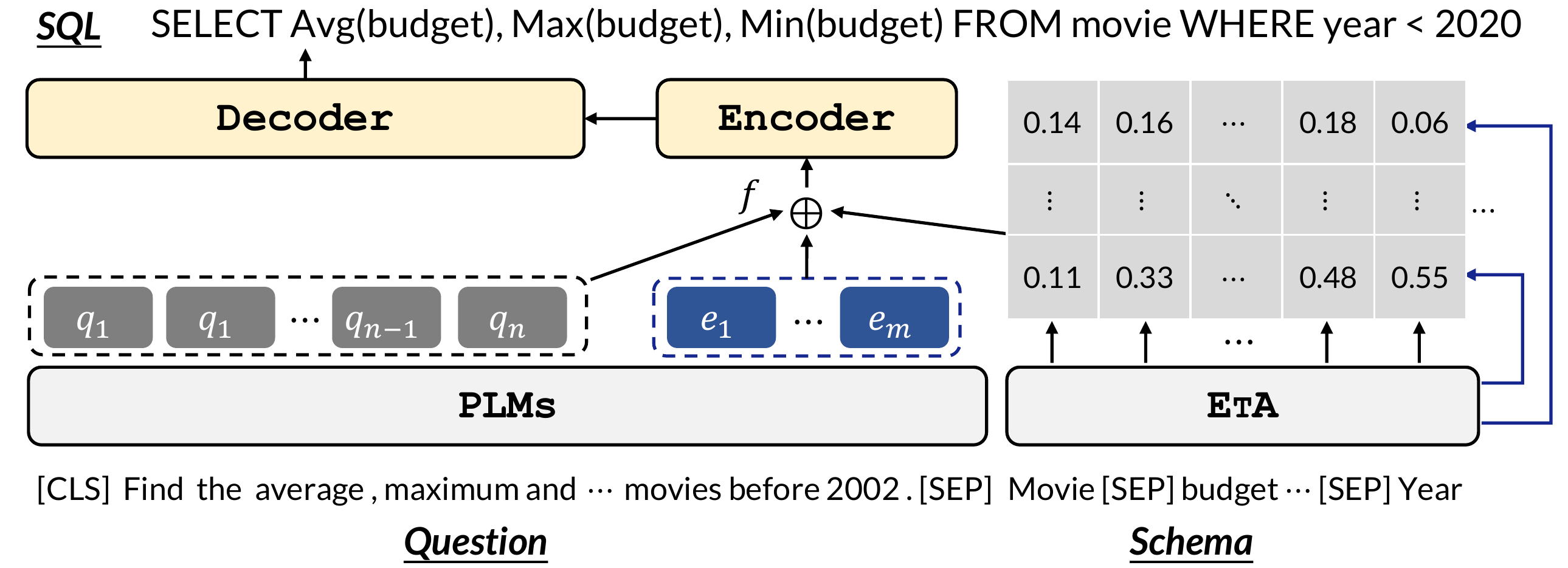}
    \caption{The illustration of the solution to couple \our with downstream text-to-SQL parsers.}
    \label{fig:pipeline_model}
\end{figure}

Inspired by \citet{lei-etal-2020-examining}, we present a general solution to couple \our with downstream parsers in \reffig{fig:pipeline_model}.
As shown, we first obtain a schema-aware representation for each question token, by fusing the token representation and its related schema representation according to the latent grounding $\boldsymbol{\alpha}{\in}\mathbb{R}^{N{\times}K}$ (gray matrix in \reffig{fig:pipeline_model}).
Specifically, given a token representation $\mathbf{q}_n$ and all schema representations $\langle \mathbf{e}_1,\mathbf{e}_2,...,\mathbf{e}_K \rangle$, the schema-aware representation $\tilde{\mathbf{q}}_n$ for $\mathbf{q}_n$ can be computed as:
\begin{equation}
    \tilde{\mathbf{q}}_n = \mathbf{q}_n {\oplus} \sum\limits_{k} \boldsymbol{\alpha}_{n,k}\,\mathbf{e}_k.
\end{equation}
Then we feed every $\tilde{\mathbf{q}}_n$ into a question encoder to generate hidden states, which are attended by a decoder to decode the SQL query.
By contributing to the schema-aware representation, \our is able to prompt the decoder to predict appropriate schemas during decoding.
Notably, the encoder and decoder are not limited to specific modules, and we follow the paper settings in subsequent experiments.

\begin{table}[t]
  \centering
  \small
  \begin{tabular}{@{\hspace{4pt}}l@{\hspace{7pt}}c@{\hspace{7pt}}c@{\hspace{7pt}}c@{\hspace{4pt}}}
  \toprule
  \multirow{2}{*}{Model} & \multicolumn{2}{c@{\hspace{3.5pt}}}{Dev} & \multicolumn{1}{c}{Test} \\
    \cmidrule(lr){2-3} \cmidrule(lr){4-4}
     & \acclf & \accexe & \accexe \\
    \toprule
    \baseline   &$37.8\pm0.6$&$56.9\pm0.7$ & $46.6\pm0.5$\\
    \baselinebert &$44.7\pm2.1$&$63.8\pm1.1$ & $51.8\pm0.4$ \\
    \ourbertpluswikiparser & $\mathbf{47.6}\pm2.5$ & $\mathbf{66.6}\pm{1.7}$ & $\mathbf{53.8}\pm0.3$ \\
    \midrule
    \alignmodel\!$^\heartsuit$ & $42.2\pm1.5$&$61.3\pm0.8$ & $49.7\pm0.4$\\
    \alignmodelbert\!$^\heartsuit$ & $47.2\pm1.2$ & $66.5\pm1.2$ & $54.1\pm0.2$ \\
    \bottomrule
  \end{tabular}
    \caption{Ex.Match and Ex.Acc results on the dev and test set of \wikiorigin.
    + BERT means using BERT to enhance encoder.
    $^\heartsuit$ means the model uses extra schema linking supervision.
    Both are the same for \reftab{tb:spider-sql-res}.
    }
  \label{tb:wikitable-sql-res}
\end{table}

\subsection{Experimental Setup}

\paragraph{Datasets and Evaluation} 

We conduct experiments on two text-to-SQL benchmarks: WikiTableQuestions(\wikiorigin) \citep{pasupat-liang-2015-compositional}\footnote{Note that the original \wikiorigin only contains answer annotations, and here we use the version with SQL annotations provided by \citet{shi-etal-2020-potential}. Our training data is a subset of the original train set, while the test data keeps the same. } and \spiderorigin \citep{yu-etal-2018-spider}.
Following previous work, we employ three kinds of evaluation metrics: \textit{Exact Match} (Ex.Match), \textit{Exact Set Match} (Ex.Set) and \textit{Execution Accuracy} (Ex.Acc).
Ex.Match evaluates the predicted SQL correctness by checking if it is equal to the ground-truth, while Ex.Set evaluates the structural correctness by checking the set match of each SQL clause in the predicted query with respect to the ground-truth.
Ex.Acc evaluates the functional correctness of the predicted SQL by checking whether it yields the ground-truth answer.

\paragraph{Baselines}

On \wikiorigin, our baselines include \baseline and {\alignmodel}, where the former is a vanilla attention based sequence to sequence model and the latter enhances \baseline with an additional schema linking task \citep{shi-etal-2020-potential}.
Similarly, on \spiderorigin, our main baselines are \baseparser and its schema linking enhanced version \slsql \citep{lei-etal-2020-examining}. \baseparser is made up of a question encoder and a two-step SQL decoder. In the first decoding step, a coarse SQL (i.e., without aggregation functions) is generated. Then the coarse SQL is used to synthesize the final SQL in the second decoding step.
Here we also report the performance of \slsqlbert (Oracle), where the learnable schema linking module is replaced with human annotations in inference.
It represents the maximum potential benefit of schema linking for the text-to-SQL task.
Meanwhile, for a comprehensive comparison, we also compare our model with state-of-the-art models on the \spiderorigin benchmark\footnote{\url{https://yale-lily.github.io/spider}}.
We refer readers to their papers for details.

\paragraph{Implementation}

As for our approach, on \wikiorigin, we employ \baseline\footnote{\url{https://github.com/tzshi/squall}} as our base parser, while on \spiderorigin we select \baseparser\footnote{\url{https://github.com/WING-NUS/slsql}} as our base parser.
For both parsers, we try to follow the same hyperparameters as described in the paper to reduce other factors that may affect the performance.
More implementation details can be found in \refsec{appendix:implement_parsing}.

\subsection{Experimental Results}

\begin{table}[t]
  \centering
  \small
  \begin{tabular}{@{\hspace{10pt}}l@{\hspace{10pt}}c@{\hspace{10pt}}c}
  \toprule 
  Model & Dev & Test \\
    \toprule
    \irnet \citep{guo-etal-2019-towards} & $61.9$ & $54.7$ \\
    \irnetsec \citep{guo-etal-2019-towards} & $63.9$ & $55.0$ \\
    \bridgelarge \citep{lin-etal-2020-bridging} & $70.0$ & $65.0$ \\
    \ratsql \citep{wang-etal-2020-rat} &  $69.7$ & $\mathbf{65.6}$ \\
    \dotline
    \baseparserbert & $57.4$ & - \\
    \baseparserlarge & $61.0$ & - \\
    \ourbertplusspiderparser & $64.5$ & $59.5$ \\
    \ourbertplusspiderparserlarge & $\mathbf{70.8}$ & $65.3$ \\
    \bottomrule
  \end{tabular}
    \caption{Ex.Set results on the dev and test set of \spiderorigin.}
  \label{tb:spider-sql-res}
\end{table}

\begin{table*}[t]
    \small
    \centering
    \begin{tabularx}{\textwidth}{p{0.40\textwidth}<{\raggedright}p{0.58\textwidth}<{\raggedright}}
    \toprule
        \multicolumn{1}{c}{\textbf{Question with Alignment}} & \multicolumn{1}{c}{\textbf{SQL with Alignment}} \\
    \midrule
        $1$. Show \redtext{name}$_1$, \redtext{country}$_2$, \redtext{age}$_3$ for all \bluetext{singers}$_4$ ordered by \redtext{age}$_3$  from the \redtext{oldest}$_3$ to the {youngest}. & SELECT \redtext{name}$_1$, \redtext{country}$_2$, \redtext{age}$_3$ FROM \bluetext{singer}$_4$ \newline ORDER BY \redtext{age}$_3$ DESC \\
    \midrule
        $2$. \redtext{Where}$_1$ is the \redtext{youngest}$_2$ \bluetext{teacher}$_3$ from? & SELECT \redtext{hometown}$_1$ FROM \bluetext{teacher}$_3$ ORDER BY \redtext{age}$_2$ ASC LIMIT 1\\
    \midrule
        $3$. {For each \bluetext{semester}$_1$, what is the \redtext{name}$_2$ and \redtext{id}$_3$ of the one with the most students \bluetext{registered}$_4$}? & SELECT \redtext{semester\_name}$_2$, \redtext{semester\_id}$_3$ FROM \bluetext{semesters}$_1$ JOIN \bluetext{student\_enrolment}$_4$ ON semesters.semester\_id = student\_enrolment.semester\_id GROUP BY \redtext{semester\_id}$_3$ \newline ORDER BY COUNT(*) DESC LIMIT 1 \\
    \bottomrule
    \end{tabularx}
    \caption{The predicted grounding pairs and SQLs of our best model on three real cases from the \spiderorigin dev set. The question token and the schema with the same subscript are grounded.}
    \label{tab:spider-examples}
\end{table*}

\reftab{tb:wikitable-sql-res} and \reftab{tb:spider-sql-res} show the experimental results of several methods on \wikiorigin and \spiderorigin respectively.
As observed, introducing \our dramatically improves the performance of both base parsers, demonstrating its effectiveness on downstream tasks.
Taking \spiderorigin as an illustration, our model \ourbertplusspiderparser boosts \baseparserbert by an absolute improvement $7.1$\% on the Ex.Set metric.
As the PLM becomes larger (e.g., BERT$_\text{L}$), the improvement becomes more significant, up to $9.8$\%.
Compared with state-of-the-art methods, our model \ourbertplusspiderparserlarge also obtains a competitive performance, which is extremely impressive since it is based on a simple parser.

More interestingly, on both datasets, our model can achieve similar even better performance compared to methods which employ extra grounding supervision.
For instance, in comparison with \slsqlbert on \spiderorigin, our \ourbertplusspiderparser outperforms it by $3.7$\%.
Taking into account that \slsql utilizes additional supervision, the performance gain is very surprising.
We attribute the gain to two possible reasons:
(1) The PLMs already learn latent grounding which is understandable to human experts.
(2) Compared with training with strong schema linking supervision, training with weak supervision alleviates the issue of exposure bias, and thus enhance the generalization ability of \our.

\begin{figure}
    \centering
    \includegraphics[width=1.0\linewidth]{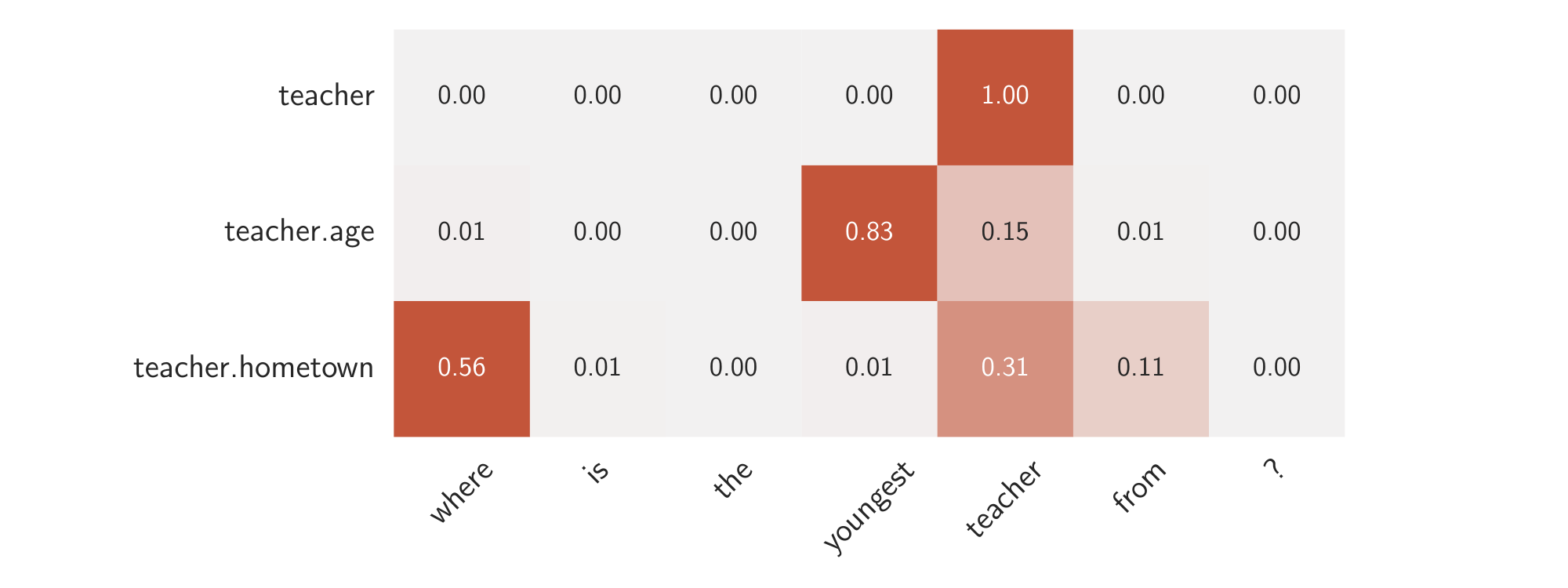}
    \caption{The latent grounding produced by \ourbertplusspiderparserlarge for the question ``\textit{Where is the youngest teacher from?}''.}
    \label{fig:vis_spider_2}
\end{figure}

\reftab{tab:spider-examples} presents the model predictions of \ourbertplusspiderparserlarge on three real cases.
As observed, \our has learned the grounding about adjective (e.g., $\text{oldest}$ ${\rightarrow}$ $\text{age}$), entity (e.g., $\text{where}$ ${\rightarrow}$ $\text{hometown}$) and semantic matching (e.g., $\text{registered}$ ${\rightarrow}$ $\text{student\_enrolment}$).
Meanwhile, grounding pairs provide us a useful guide to better understand the model predictions.
\reffig{fig:vis_spider_2} visualizes the latent grounding for Q$2$ in \reftab{tab:spider-examples}, and more visualization can be found in \refsec{appendix:vis}.

\section{Related Work}

The most related work to ours is the line of inducing or probing knowledge in pretrained language models.
According to the knowledge category, there are mainly two kinds of methods: one focuses on syntactic knowledge and the other pays attention to semantic knowledge.
Under the category of syntactic knowledge, several work showed that BERT embeddings encoded syntactic information in a structural form and can be recovered \citep{lin-etal-2019-open, warstadt2020neural, hewitt-manning-2019-structural, wu-etal-2020-perturbed}.
However, recent work also showed that BERT did not rely on syntactic information for downstream task performance, and thus doubted the role of syntactic knowledge \citep{ettinger-2020-bert, glavas2020supervised}.
As for semantic knowledge, although it is less explored than syntactic knowledge, previous work showed that BERT contained some semantic information, such as entity types \citep{ettinger-2020-bert}, semantic roles \citep{tenney2018you} and factual knowledge \citep{petroni-etal-2019-language}.
Different from the above work, we focus on the grounding capability, an under-explored branch of language semantics.

Our work is also closely related to entity linking and schema linking, which can be viewed as subareas of grounding on specific scenarios.
Given an utterance, entity linking aims at finding all mentioned entities in it using a knowledge base as candidate pool \citep{tan-etal-2017-entity,entity-linking-with-fine-grained, li-etal-2020-efficient}, while schema linking tries to find all mentioned schemas related to specific databases \citep{dong-etal-2019-data,lei-etal-2020-examining,shi-etal-2020-potential}.
Previous work generally either employed full supervision to train linking models \citep{li-etal-2020-efficient,lei-etal-2020-examining,shi-etal-2020-potential}, or treated linking as a minor pre-processing\citep{yu-etal-2018-typesql, guo-etal-2019-towards,DBLP:gramma-based} and used heuristic rules to obtain the result.
Our work is different from them since we optimize the linking model with weak supervision from downstream signals, which is flexible and practicable.
Similarly, \citet{dong-etal-2019-data} utilized downstream supervision to train their linking model.
Compared with them using policy gradient, our method is more efficient since it directly learns the grounding module using pseudo alignment as supervision.

\section{Conclusion \& Future Work}

In summary, we propose a novel weakly supervised approach to awaken latent grounding from pretrained language models via erasing.
Only with downstream signals, our approach can induce latent grounding from pretrained language models which is understandable to human experts.
More importantly, we demonstrate that our approach could be applied to off-the-shelf text-to-SQL parsers and significantly improve their performance.
For future work, we plan to extend our approach to more downstream tasks such as visual question answering.
We also plan to utilize our approach to improve the error locator module in existing interactive semantic parsing systems \citep{li-etal-2020-mean}.

\section*{Acknowledgement}

We would like to thank all the anonymous reviewers for their constructive feedback and useful comments.
We also thank Tao Yu and Bo Pang for evaluating our submitted models on the test set of Spider.
The first author Qian is supported by the Academic Excellence Foundation of Beihang University for PhD Students.

\section*{Ethical Considerations}
This paper conducts experiments on several existing datasets covering the areas of entity linking, schema entity and text-to-SQL. All claims in this paper are based on the experimental results. Every experiment can be conducted on a single Tesla P100 or P40 GPU. No demographic or identity characteristics information is used in this paper. 

\bibliographystyle{acl_natbib}
\bibliography{anthology,acl2021}

\clearpage
\appendix

\section{Evaluation Details}\label{appendix:evaluation}
\begin{table*}[t]
\centering
\begin{tabular}{ccccccc}
\toprule
\multirow{2}{*}{Dataset} & \multicolumn{2}{c}{Train} & \multicolumn{2}{c}{Dev} & \multicolumn{2}{c}{Test} \\ \cmidrule(lr){2-3} \cmidrule(lr){4-5} \cmidrule(lr){6-7} 
                         & \#Q          & \#C        & \#Q         & \#C       & \#Q         & \#C        \\ \midrule
\wikidata                   & $9,030$         & $19,185$         & $2,246$        & $4,774$        & --       & --         \\
\spiderdata                & $7,000$         & $28,848$         & $1,034$        & $4,360$       & --          & --         \\
\wikiorigin                  & $9,030$         & --         & $2,246$        & --        & $4,344$        & --         \\
\spiderorigin                   & $7,000$         & --         & $1,034$        & --        & $2,147$        & --         \\
\webqspdata                   & $2,974$         & $3,242$       & --          & --        & $1,603$        & $1,806$       \\
\graphquesdata                 & $2,089$         & $2,253$       & --          & --        & $2,075$        & $2,229$       \\ \bottomrule
\end{tabular}
\caption{Statistics for all datasets used in our experiments. For \wikidata and \wikiorigin, we only show the size of Split-0, and details of other splits can be found in Table \ref{tb:squall_split_size}. \#Q represents the number of questions, \#C represents the number of concepts.}
\label{tb:dataset_details}
\end{table*}

\begin{table}[t]
  \centering
  \begin{tabular}{ccc}
  \toprule 
  Split & Train & Dev \\
    \midrule
    $0$ & $9,030$ & $2,246$ \\
    $1$ & $9,032$ & $2,244$ \\
    $2$ & $9,028$ & $2,248$ \\
    $3$ & $8,945$ & $2,331$ \\
    $4$ & $9,069$ & $2,207$ \\
    \bottomrule
  \end{tabular}
    \caption{The size of train set and dev set of five splits on {\wikidata} and \wikiorigin.}
  \label{tb:squall_split_size}
\end{table}

\subsection{Schema Linking}

Let $\Omega_{col}$ be a set $\{(c,q)_i|1\leq i\leq N\}$ which contains $N$ gold (column-question token) tuples.
Let $\overline{\Omega}_{col}$ be a set $\{(\overline{c},\overline{q})_j|1\leq j\leq M\}$ which contains $M$ predicted (column-question token) tuples.
We define the precision(\colprec), recall(\colrecall), F1-score(\colfscore) as:
\[
\frac{\left|\Gamma_{col}\right|}{\left|\overline{\Omega}_{col}\right|}, \frac{\left|\Gamma_{col}\right|}{\left|{\Omega}_{col}\right|}, \frac{2\text{Col}_P \text{Col}_R}{\text{Col}_P + \text{Col}_R}
\]
where ${\Gamma}_{col} = {\Omega}_{col} \bigcap {\overline{\Omega}_{col}}$.
The definitions of \tabprec, \tabrecall, \tabfscore are similar.
Note that the result reported in Table $8$ of \citet{shi-etal-2020-potential} use a different evaluation metrics. Here we re-evaluate their model by the above mentioned metrics for fair comparison.

\subsection{Entity Linking}

Let $\Omega=\{(e,[q_s,q_e])_i|1\leq i\leq N\}$ be the gold entity-mention set and $\overline{\Omega}=\{(\overline{e},[\overline{q_s},\overline{q_e}])_j|1\leq j\leq M\}$ be the predicted entity-mention set, where $e$ is the entity, $q_e,q_s$ are the mention boundaries in the question $q$.
In the weak matching setting, a prediction is correct only if the ground-truth entity is identified and the predicted mention boundaries overlap with the ground-truth boundaries. Therefore, the True-Positive prediction set is defined as:
\begin{align*}
    \Gamma = \{e|(e,[q_s,q_e])\in \Omega, (e,[\overline{q_s},\overline{q_e}]) \in \overline{\Omega}, \\ [q_s,q_e] \bigcap [\overline{q_s},\overline{q_e}] \neq \emptyset \}.
\end{align*}
The corresponding precision(\entprec), recall(\entrecall) and F1(\entfscore) are:
\[\frac{\left|\Gamma\right|}{\left|\overline{\Omega}\right|}, \frac{\left|\Gamma\right|}{\left|{\Omega}\right|}, \frac{2\text{Ent}_P \text{Ent}_R}{\text{Ent}_P + \text{Ent}_R}\]

\section{Dataset Statistic}

All details of datasets used in this paper are shown in Table \ref{tb:dataset_details}.

\section{Implementation Details}

For all experiments, we employ the AdamW optimizer and the default learning rate schedule strategy provided by Transformers library \citep{wolf-etal-2020-transformers}.

\subsection{Experiments on Grounding}\label{appendix:implement_grounding}

\paragraph{\wikidata}

We use uncased BERT-base as the encoder.
The learning rate is $3\times10^{-5}$.
The training epoch is $50$ with a batch size of $16$.
The dropout rate and the threshold $\tau$ are set to $0.3$ and $0.2$ respectively.
The training process lasts $6$ hours on a single 16GB Tesla P100 GPU.

\paragraph{\spiderdata}

We implement two versions: uncased BERT-base and uncased BERT-large.
For both versions, the learning rate is $5\times10^{-5}$ and the training epoch is $50$.
For BERT-base (BERT-large) version, the batch size and gradient accumulation step are set to $12$ ($6$) and $6$ ($4$).
The dropout rate and the threshold $\tau$ are set to $0.3$ and $0.2$ respectively.
As for training time, BERT-base (BERT-large) version is trained on a 24GB Tesla P40 and it takes about $16$ ($48$) hours to finish the training process.

\paragraph{\webqspdata \& \graphquesdata}
Due to the large amount of entity candidates, we first use the candidate retrieval method proposed in \citep{sorokin-gurevych-2018-mixing} to reduce the number of candidates.
After that, we still can not feed all candidates along with the question due to the maximum encoding length of BERT.
Therefore, we divide the candidates into multiple chunks and feed each chunk (along with the question) into BERT sequentially. 

In implementation, we use uncased BERT-base as the encoder.
The learning rate is $1\times10^{-5}$
The training epoch is $50$ with a batch size of $16$.
The dropout rate and the threshold $\tau$ are set to $0.3$ and $0.3$ respectively.
The training procedure finishes within $10$ hours on a single Tesla M40 GPU.

\subsection{Experiments on Text-to-SQL}\label{appendix:implement_parsing}

\begin{table}[t]
  \centering
  \begin{tabular}{lccc}
  \toprule
  \multirow{2}{*}{Split} & \multicolumn{2}{c@{\hspace{3.5pt}}}{Dev} & \multicolumn{1}{c}{Test} \\
    \cmidrule(lr){2-3} \cmidrule(lr){4-4}
     & \acclf & \accexe & \accexe \\
    \midrule
    $0$   
    &$45.10$ & $64.43$ & $53.57$\\
    $1$
    &$47.39$ & $67.01$ & $54.17$ \\
    $2$
    & $47.24$ & $65.93$ & $53.61$ \\
    
    $3$
    & $45.99$ & $65.72$ & $53.41$\\
    $4$
    & $52.38$ & $69.73$ & $52.41$ \\
    \bottomrule
  \end{tabular}
    \caption{The experimental results of all splits on \wikiorigin.}
  \label{tb:wikitable-sql-five-splits-res}
\end{table}

For experiments of the text-to-SQL task, we employ the official code released along with \citet{shi-etal-2020-potential} (on \wikiorigin) and \citet{lei-etal-2020-examining} (on \spiderorigin).
When coupling \our with these models, we first produce a one-hot grounding matrix derived by grounding pairs and then feed it into them as described in \refsec{sec:case_study}.

\paragraph{\wikiorigin}

We use uncased BERT-base as the encoder.
The training epoch is $50$ with a batch size of $8$.
The learning rate is $1 \times 10^{-5}$ for the BERT module and $1 \times 10^{-3}$ for other modules.
The dropout rate is set to $0.2$.
The training process finishes within $16$ hours on a single 16GB Tesla P100 GPU.

Meanwhile, we follow the previous work \citep{shi-etal-2020-potential} to employ $5$-fold cross-validation, and experimental results of all five splits on \wikiorigin using \ourbertpluswikiparser are shown in  \reftab{tb:wikitable-sql-five-splits-res}.

\paragraph{\spiderorigin}
  
We implement two versions: uncased BERT-base and uncased BERT-large.
For BERT-base (BERT-large), the learning rate is $1.25\times10^{-5}$ ($6.25\times10^{-6}$) for the BERT module and $1\times10^{-4}$ ($5\times10^{-5}$) for other modules.
The batch size and gradient accumulation step are set to $10$ ($6$) and $5$ ($4$) for BERT-base (BERT-large) version.
The dropout rate is set to $0.3$.
As for training time, BERT-base (BERT-large) version is trained on a 24GB Tesla P40 and it takes about $36$ ($56$) hours to finish the training process.

\section{Latent Grounding Visualization}\label{appendix:vis}
\begin{figure*}
    \centering
    \includegraphics[width=1.0\linewidth]{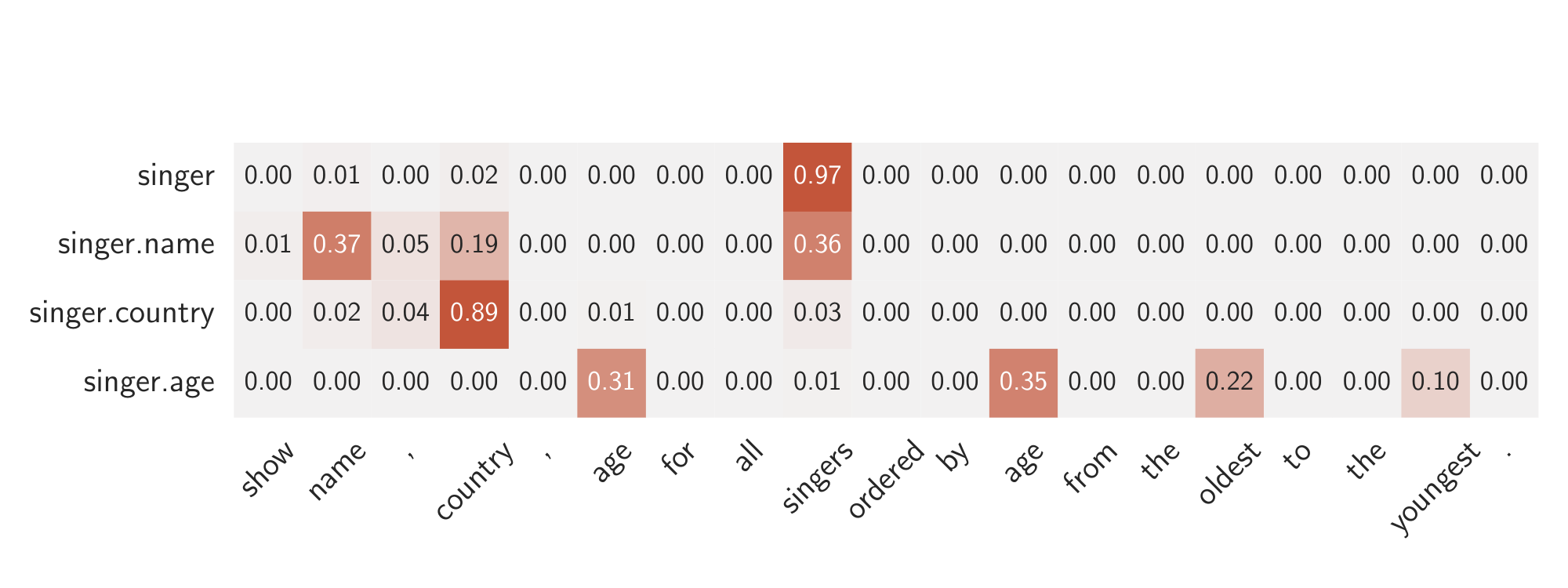}
    \caption{The latent grounding produced by \ourbertplusspiderparserlarge for the question ``\textit{Show name, country, age for all singers ordered by age from the oldest to the youngest.}''.}
    \label{fig:vis_spider_1}
\end{figure*}

\begin{figure*}
    \centering
    \includegraphics[width=1.0\linewidth]{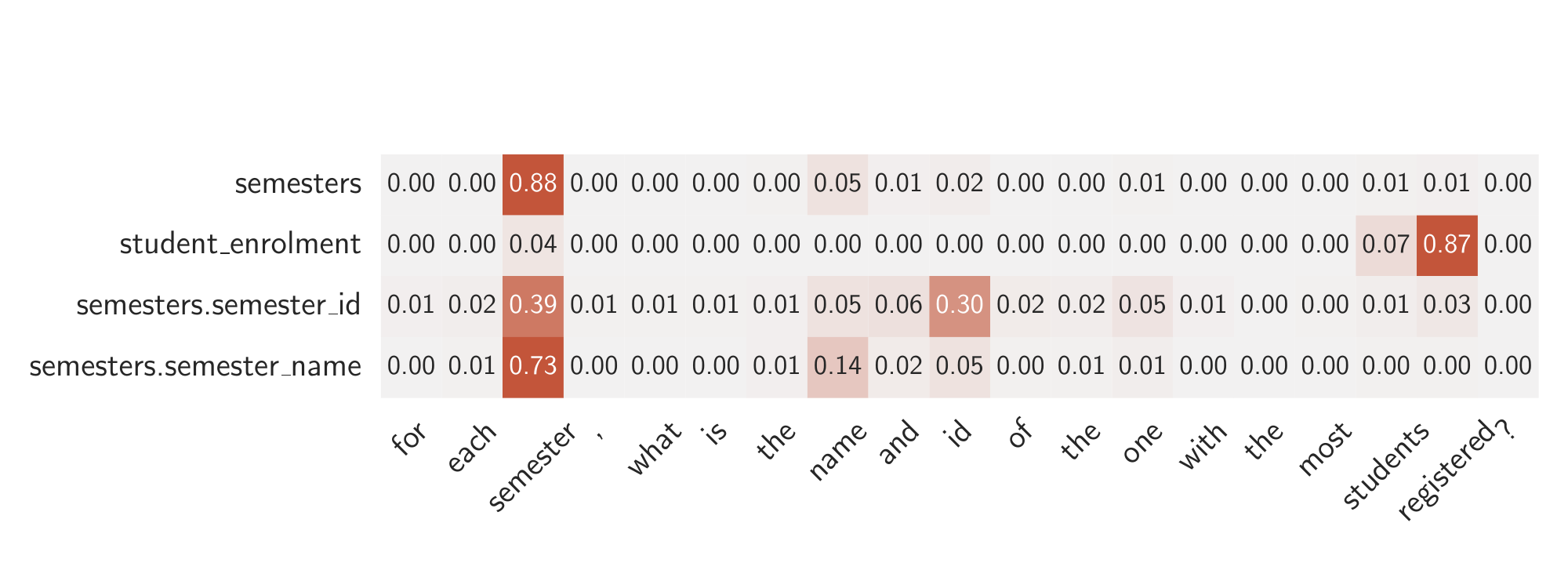}
    \caption{The latent grounding produced by \ourbertplusspiderparserlarge for the question ``\textit{For each semester, what is the name and id of the one with the most students registered?}''.}
    \label{fig:vis_spider_3}
\end{figure*}

\reffig{fig:vis_spider_1} and \reffig{fig:vis_spider_3} show the latent grounding visualization corresponding to examples in \reftab{tab:spider-examples}.


\end{document}

%% file: tk_plot.tex
\begin{figure}[t!]
\begin{center}
\begin{tikzpicture}[scale=0.85]
    \begin{axis}[
        legend style={nodes={scale=0.55, transform shape}},
        xmin=0, xmax=25,
        ymin=0.0, ymax=75.0,
        xtick={0,5,10,15,20,25},
        ytick={15,30,45,60,75},
        legend pos=south east,
        ymajorgrids=true,
        grid style=dashed,
    ]
	\addplot[
        color=black,
        mark=square,
        mark size=1pt
	] coordinates {
        (1,65.1)
        (4,67.6)
        (7,67.6)
        (10,68.3)
        (13,68.1)
        (16,68.8)
        (19,68.3)
        (22,68)
        (24,68.3)
    };
	\addlegendentry{Awakening}

    \addplot[
        color=purple,
        mark=square,
        mark size=1pt
    ]
    coordinates {
        (1,38.5)
        (4,33.6)
        (7,35.7)
        (10,36.1)
        (13,36.6)
        (16,34.3)
        (19,34.7)
        (22,37.3)
        (24,33.9)
    };
    \addlegendentry{Pseudo Alignment}

	\addplot[
        color=blue,
        mark=square,
        mark size=1pt
	]coordinates {
        (1,35.7)
        (4,34.5)
        (7,36.6)
        (10,37.2)
        (13,36.6)
        (16,35.8)
        (19,38)
        (22,35.8)
        (24,36.4)
    };
	\addlegendentry{Pseudo w/ $\mathtt{Softmax}$}

	\addplot[
        color=orange,
        mark=square,
        mark size=1pt
	]
	coordinates {
        (1,57.7)
        (4,58.2)
        (7,57.8)
        (10,54.6)
        (13,55.3)
        (16,57.9)
        (19,51.2)
        (22,56)
        (24,53)
    };
	\addlegendentry{Pseudo w/ $\mathtt{Sum}$}
	\end{axis}
\end{tikzpicture}
\end{center}
\caption{\colfscore score on the dev set of \wikidata at different training epochs. ``Pseudo w/ $\mathtt{Softmax}$'' means normalizing pseudo alignment with $\mathtt{Softmax}$, while ``Pseudo w/ $\mathtt{Sum}$'' means normalizing through dividing each number by the sum of them.}
\label{fig:self_teaching}
\vspace{-2mm}
\end{figure}

%% file: main.bbl
\begin{thebibliography}{49}
\expandafter\ifx\csname natexlab\endcsname\relax\def\natexlab#1{#1}\fi

\bibitem[{Arras et~al.(2016)Arras, Horn, Montavon, M{\"{u}}ller, and
  Samek}]{DBLP:journals/corr/ArrasHMMS16a}
Leila Arras, Franziska Horn, Gr{\'{e}}goire Montavon, Klaus{-}Robert
  M{\"{u}}ller, and Wojciech Samek. 2016.
\newblock \href {http://arxiv.org/abs/1612.07843} {"what is relevant in a text
  document?": An interpretable machine learning approach}.
\newblock \emph{CoRR}, abs/1612.07843.

\bibitem[{Brown et~al.(2020)Brown, Mann, Ryder, Subbiah, Kaplan, Dhariwal,
  Neelakantan, Shyam, Sastry, Askell, Agarwal, Herbert{-}Voss, Krueger,
  Henighan, Child, Ramesh, Ziegler, Wu, Winter, Hesse, Chen, Sigler, Litwin,
  Gray, Chess, Clark, Berner, McCandlish, Radford, Sutskever, and
  Amodei}]{brown2020language}
Tom~B. Brown, Benjamin Mann, Nick Ryder, Melanie Subbiah, Jared Kaplan,
  Prafulla Dhariwal, Arvind Neelakantan, Pranav Shyam, Girish Sastry, Amanda
  Askell, Sandhini Agarwal, Ariel Herbert{-}Voss, Gretchen Krueger, Tom
  Henighan, Rewon Child, Aditya Ramesh, Daniel~M. Ziegler, Jeffrey Wu, Clemens
  Winter, Christopher Hesse, Mark Chen, Eric Sigler, Mateusz Litwin, Scott
  Gray, Benjamin Chess, Jack Clark, Christopher Berner, Sam McCandlish, Alec
  Radford, Ilya Sutskever, and Dario Amodei. 2020.
\newblock \href
  {https://proceedings.neurips.cc/paper/2020/hash/1457c0d6bfcb4967418bfb8ac142f64a-Abstract.html}
  {Language models are few-shot learners}.
\newblock In \emph{Advances in Neural Information Processing Systems 33: Annual
  Conference on Neural Information Processing Systems 2020, NeurIPS 2020,
  December 6-12, 2020, virtual}.

\bibitem[{Chen et~al.(2018)Chen, Liang, Xie, and
  Xiao}]{entity-linking-with-fine-grained}
Lihan Chen, Jiaqing Liang, Chenhao Xie, and Yanghua Xiao. 2018.
\newblock \href {https://doi.org/10.1145/3269206.3271809} {Short text entity
  linking with fine-grained topics}.
\newblock In \emph{Proceedings of the 27th ACM International Conference on
  Information and Knowledge Management}, CIKM '18, page 457–466, New York,
  NY, USA. Association for Computing Machinery.

\bibitem[{Chen et~al.(2020)Chen, San, Liu, and Ji}]{chen-etal-2020-tale}
Sanxing Chen, Aidan San, Xiaodong Liu, and Yangfeng Ji. 2020.
\newblock \href {https://www.aclweb.org/anthology/2020.coling-main.260} {A tale
  of two linkings: Dynamically gating between schema linking and structural
  linking for text-to-{SQL} parsing}.
\newblock In \emph{Proceedings of the 28th International Conference on
  Computational Linguistics}, pages 2900--2912, Barcelona, Spain (Online).
  International Committee on Computational Linguistics.

\bibitem[{Cheng et~al.(2017)Cheng, Reddy, Saraswat, and
  Lapata}]{cheng-etal-2017-learning}
Jianpeng Cheng, Siva Reddy, Vijay Saraswat, and Mirella Lapata. 2017.
\newblock \href {https://doi.org/10.18653/v1/P17-1005} {Learning structured
  natural language representations for semantic parsing}.
\newblock In \emph{Proceedings of the 55th Annual Meeting of the Association
  for Computational Linguistics (Volume 1: Long Papers)}, pages 44--55,
  Vancouver, Canada. Association for Computational Linguistics.

\bibitem[{Devlin et~al.(2019)Devlin, Chang, Lee, and
  Toutanova}]{devlin-etal-2019-bert}
Jacob Devlin, Ming-Wei Chang, Kenton Lee, and Kristina Toutanova. 2019.
\newblock \href {https://doi.org/10.18653/v1/N19-1423} {{BERT}: Pre-training of
  deep bidirectional transformers for language understanding}.
\newblock In \emph{Proceedings of the 2019 Conference of the North {A}merican
  Chapter of the Association for Computational Linguistics: Human Language
  Technologies, Volume 1 (Long and Short Papers)}, pages 4171--4186,
  Minneapolis, Minnesota. Association for Computational Linguistics.

\bibitem[{Dong et~al.(2019)Dong, Sun, Liu, Lou, and
  Zhang}]{dong-etal-2019-data}
Zhen Dong, Shizhao Sun, Hongzhi Liu, Jian-Guang Lou, and Dongmei Zhang. 2019.
\newblock \href {https://doi.org/10.18653/v1/D19-1543} {Data-anonymous encoding
  for text-to-{SQL} generation}.
\newblock In \emph{Proceedings of the 2019 Conference on Empirical Methods in
  Natural Language Processing and the 9th International Joint Conference on
  Natural Language Processing (EMNLP-IJCNLP)}, pages 5405--5414, Hong Kong,
  China. Association for Computational Linguistics.

\bibitem[{Ettinger(2020)}]{ettinger-2020-bert}
Allyson Ettinger. 2020.
\newblock \href {https://doi.org/10.1162/tacl_a_00298} {What {BERT} is not:
  Lessons from a new suite of psycholinguistic diagnostics for language
  models}.
\newblock \emph{Transactions of the Association for Computational Linguistics},
  8:34--48.

\bibitem[{Glavas and Vulic(2020)}]{glavas2020supervised}
Goran Glavas and Ivan Vulic. 2020.
\newblock \href {http://arxiv.org/abs/2008.06788} {Is supervised syntactic
  parsing beneficial for language understanding? an empirical investigation}.
\newblock \emph{CoRR}, abs/2008.06788.

\bibitem[{Guo et~al.(2019)Guo, Zhan, Gao, Xiao, Lou, Liu, and
  Zhang}]{guo-etal-2019-towards}
Jiaqi Guo, Zecheng Zhan, Yan Gao, Yan Xiao, Jian-Guang Lou, Ting Liu, and
  Dongmei Zhang. 2019.
\newblock \href {https://doi.org/10.18653/v1/P19-1444} {Towards complex
  text-to-{SQL} in cross-domain database with intermediate representation}.
\newblock In \emph{Proceedings of the 57th Annual Meeting of the Association
  for Computational Linguistics}, pages 4524--4535, Florence, Italy.
  Association for Computational Linguistics.

\bibitem[{Hewitt and Liang(2019)}]{hewitt-liang-2019-designing}
John Hewitt and Percy Liang. 2019.
\newblock \href {https://doi.org/10.18653/v1/D19-1275} {Designing and
  interpreting probes with control tasks}.
\newblock In \emph{Proceedings of the 2019 Conference on Empirical Methods in
  Natural Language Processing and the 9th International Joint Conference on
  Natural Language Processing (EMNLP-IJCNLP)}, pages 2733--2743, Hong Kong,
  China. Association for Computational Linguistics.

\bibitem[{Hewitt and Manning(2019)}]{hewitt-manning-2019-structural}
John Hewitt and Christopher~D. Manning. 2019.
\newblock \href {https://doi.org/10.18653/v1/N19-1419} {{A} structural probe
  for finding syntax in word representations}.
\newblock In \emph{Proceedings of the 2019 Conference of the North {A}merican
  Chapter of the Association for Computational Linguistics: Human Language
  Technologies, Volume 1 (Long and Short Papers)}, pages 4129--4138,
  Minneapolis, Minnesota. Association for Computational Linguistics.

\bibitem[{Hwang et~al.(2019)Hwang, Yim, Park, and Seo}]{Hwang2019ACE}
Wonseok Hwang, Jinyeung Yim, Seunghyun Park, and Minjoon Seo. 2019.
\newblock \href {http://arxiv.org/abs/1902.01069} {A comprehensive exploration
  on wikisql with table-aware word contextualization}.
\newblock \emph{CoRR}, abs/1902.01069.

\bibitem[{Jawahar et~al.(2019)Jawahar, Sagot, and
  Seddah}]{jawahar-etal-2019-bert}
Ganesh Jawahar, Beno{\^\i}t Sagot, and Djam{\'e} Seddah. 2019.
\newblock \href {https://doi.org/10.18653/v1/P19-1356} {What does {BERT} learn
  about the structure of language?}
\newblock In \emph{Proceedings of the 57th Annual Meeting of the Association
  for Computational Linguistics}, pages 3651--3657, Florence, Italy.
  Association for Computational Linguistics.

\bibitem[{Lei et~al.(2020)Lei, Wang, Ma, Gan, Lu, Kan, and
  Chua}]{lei-etal-2020-examining}
Wenqiang Lei, Weixin Wang, Zhixin Ma, Tian Gan, Wei Lu, Min-Yen Kan, and
  Tat-Seng Chua. 2020.
\newblock \href {https://doi.org/10.18653/v1/2020.emnlp-main.564} {Re-examining
  the role of schema linking in text-to-{SQL}}.
\newblock In \emph{Proceedings of the 2020 Conference on Empirical Methods in
  Natural Language Processing (EMNLP)}, pages 6943--6954, Online. Association
  for Computational Linguistics.

\bibitem[{Li et~al.(2020{\natexlab{a}})Li, Min, Iyer, Mehdad, and
  Yih}]{li-etal-2020-efficient}
Belinda~Z. Li, Sewon Min, Srinivasan Iyer, Yashar Mehdad, and Wen-tau Yih.
  2020{\natexlab{a}}.
\newblock \href {https://doi.org/10.18653/v1/2020.emnlp-main.522} {Efficient
  one-pass end-to-end entity linking for questions}.
\newblock In \emph{Proceedings of the 2020 Conference on Empirical Methods in
  Natural Language Processing (EMNLP)}, pages 6433--6441, Online. Association
  for Computational Linguistics.

\bibitem[{Li et~al.(2020{\natexlab{b}})Li, Chen, Liu, Gao, Lou, Zhang, and
  Zhang}]{li-etal-2020-mean}
Yuntao Li, Bei Chen, Qian Liu, Yan Gao, Jian-Guang Lou, Yan Zhang, and Dongmei
  Zhang. 2020{\natexlab{b}}.
\newblock \href {https://doi.org/10.18653/v1/2020.emnlp-main.561} {{``}what do
  you mean by that?{''} a parser-independent interactive approach for enhancing
  text-to-{SQL}}.
\newblock In \emph{Proceedings of the 2020 Conference on Empirical Methods in
  Natural Language Processing (EMNLP)}, pages 6913--6922, Online. Association
  for Computational Linguistics.

\bibitem[{Liang et~al.(2013)Liang, Jordan, and
  Klein}]{liang-etal-2013-learning}
Percy Liang, Michael~I. Jordan, and Dan Klein. 2013.
\newblock \href {https://doi.org/10.1162/COLI_a_00127} {Learning
  dependency-based compositional semantics}.
\newblock \emph{Computational Linguistics}, 39(2):389--446.

\bibitem[{Lin et~al.(2019{\natexlab{a}})Lin, Bogin, Neumann, Berant, and
  Gardner}]{DBLP:gramma-based}
Kevin Lin, Ben Bogin, Mark Neumann, Jonathan Berant, and Matt Gardner.
  2019{\natexlab{a}}.
\newblock \href {http://arxiv.org/abs/1905.13326} {Grammar-based neural
  text-to-sql generation}.
\newblock \emph{CoRR}, abs/1905.13326.

\bibitem[{Lin et~al.(2020)Lin, Socher, and Xiong}]{lin-etal-2020-bridging}
Xi~Victoria Lin, Richard Socher, and Caiming Xiong. 2020.
\newblock \href {https://doi.org/10.18653/v1/2020.findings-emnlp.438} {Bridging
  textual and tabular data for cross-domain text-to-{SQL} semantic parsing}.
\newblock In \emph{Findings of the Association for Computational Linguistics:
  EMNLP 2020}, pages 4870--4888, Online. Association for Computational
  Linguistics.

\bibitem[{Lin et~al.(2019{\natexlab{b}})Lin, Tan, and
  Frank}]{lin-etal-2019-open}
Yongjie Lin, Yi~Chern Tan, and Robert Frank. 2019{\natexlab{b}}.
\newblock \href {https://doi.org/10.18653/v1/W19-4825} {Open sesame: Getting
  inside {BERT}{'}s linguistic knowledge}.
\newblock In \emph{Proceedings of the 2019 ACL Workshop BlackboxNLP: Analyzing
  and Interpreting Neural Networks for NLP}, pages 241--253, Florence, Italy.
  Association for Computational Linguistics.

\bibitem[{Liu et~al.(2019)Liu, Gardner, Belinkov, Peters, and
  Smith}]{liu-etal-2019-linguistic}
Nelson~F. Liu, Matt Gardner, Yonatan Belinkov, Matthew~E. Peters, and Noah~A.
  Smith. 2019.
\newblock \href {https://doi.org/10.18653/v1/N19-1112} {Linguistic knowledge
  and transferability of contextual representations}.
\newblock In \emph{Proceedings of the 2019 Conference of the North {A}merican
  Chapter of the Association for Computational Linguistics: Human Language
  Technologies, Volume 1 (Long and Short Papers)}, pages 1073--1094,
  Minneapolis, Minnesota. Association for Computational Linguistics.

\bibitem[{Liu et~al.(2020{\natexlab{a}})Liu, Chen, Guo, Lou, Zhou, and
  Zhang}]{qian2020how}
Qian Liu, Bei Chen, Jiaqi Guo, Jian-Guang Lou, Bin Zhou, and Dongmei Zhang.
  2020{\natexlab{a}}.
\newblock \href {https://arxiv.org/pdf/2002.00652.pdf} {How far are we from
  effective context modeling? an exploratory study on semantic parsing in
  context twitter}.
\newblock In \emph{IJCAI}, pages 3580--3586.

\bibitem[{Liu et~al.(2020{\natexlab{b}})Liu, Chen, Chen, Lou, Chen, Zhou, and
  Zhang}]{liu-etal-2020-impress}
Qian Liu, Yihong Chen, Bei Chen, Jian-Guang Lou, Zixuan Chen, Bin Zhou, and
  Dongmei Zhang. 2020{\natexlab{b}}.
\newblock \href {https://doi.org/10.18653/v1/2020.acl-main.131} {You impress
  me: Dialogue generation via mutual persona perception}.
\newblock In \emph{Proceedings of the 58th Annual Meeting of the Association
  for Computational Linguistics}, pages 1417--1427, Online. Association for
  Computational Linguistics.

\bibitem[{Loshchilov and Hutter(2019)}]{loshchilov2018decoupled}
Ilya Loshchilov and Frank Hutter. 2019.
\newblock \href {https://openreview.net/forum?id=Bkg6RiCqY7} {Decoupled weight
  decay regularization}.
\newblock In \emph{7th International Conference on Learning Representations,
  {ICLR} 2019, New Orleans, LA, USA, May 6-9, 2019}. OpenReview.net.

\bibitem[{Pasupat and Liang(2015)}]{pasupat-liang-2015-compositional}
Panupong Pasupat and Percy Liang. 2015.
\newblock \href {https://doi.org/10.3115/v1/P15-1142} {Compositional semantic
  parsing on semi-structured tables}.
\newblock In \emph{Proceedings of the 53rd Annual Meeting of the Association
  for Computational Linguistics and the 7th International Joint Conference on
  Natural Language Processing (Volume 1: Long Papers)}, pages 1470--1480,
  Beijing, China. Association for Computational Linguistics.

\bibitem[{Paszke et~al.(2019)Paszke, Gross, Massa, Lerer, Bradbury, Chanan,
  Killeen, Lin, Gimelshein, Antiga, Desmaison, Kopf, Yang, DeVito, Raison,
  Tejani, Chilamkurthy, Steiner, Fang, Bai, and
  Chintala}]{NEURIPS2019_bdbca288}
Adam Paszke, Sam Gross, Francisco Massa, Adam Lerer, James Bradbury, Gregory
  Chanan, Trevor Killeen, Zeming Lin, Natalia Gimelshein, Luca Antiga, Alban
  Desmaison, Andreas Kopf, Edward Yang, Zachary DeVito, Martin Raison, Alykhan
  Tejani, Sasank Chilamkurthy, Benoit Steiner, Lu~Fang, Junjie Bai, and Soumith
  Chintala. 2019.
\newblock \href
  {https://proceedings.neurips.cc/paper/2019/file/bdbca288fee7f92f2bfa9f7012727740-Paper.pdf}
  {Pytorch: An imperative style, high-performance deep learning library}.
\newblock In \emph{Advances in Neural Information Processing Systems},
  volume~32, pages 8026--8037. Curran Associates, Inc.

\bibitem[{Petroni et~al.(2019)Petroni, Rockt{\"a}schel, Riedel, Lewis, Bakhtin,
  Wu, and Miller}]{petroni-etal-2019-language}
Fabio Petroni, Tim Rockt{\"a}schel, Sebastian Riedel, Patrick Lewis, Anton
  Bakhtin, Yuxiang Wu, and Alexander Miller. 2019.
\newblock \href {https://doi.org/10.18653/v1/D19-1250} {Language models as
  knowledge bases?}
\newblock In \emph{Proceedings of the 2019 Conference on Empirical Methods in
  Natural Language Processing and the 9th International Joint Conference on
  Natural Language Processing (EMNLP-IJCNLP)}, pages 2463--2473, Hong Kong,
  China. Association for Computational Linguistics.

\bibitem[{Reddy et~al.(2016)Reddy, T{\"a}ckstr{\"o}m, Collins, Kwiatkowski,
  Das, Steedman, and Lapata}]{reddy-etal-2016-transforming}
Siva Reddy, Oscar T{\"a}ckstr{\"o}m, Michael Collins, Tom Kwiatkowski, Dipanjan
  Das, Mark Steedman, and Mirella Lapata. 2016.
\newblock \href {https://doi.org/10.1162/tacl_a_00088} {Transforming dependency
  structures to logical forms for semantic parsing}.
\newblock \emph{Transactions of the Association for Computational Linguistics},
  4:127--140.

\bibitem[{Rogers et~al.(2020)Rogers, Kovaleva, and
  Rumshisky}]{anna-etal-2021-primer}
Anna Rogers, Olga Kovaleva, and Anna Rumshisky. 2020.
\newblock \href {https://doi.org/10.1162/tacl\_a\_00349} {A primer in
  bertology: What we know about how bert works}.
\newblock \emph{Transactions of the Association for Computational Linguistics},
  8:842--866.

\bibitem[{Roy(2005)}]{roy2005grounding}
Deb Roy. 2005.
\newblock \href {https://doi.org/https://doi.org/10.1016/j.tics.2005.06.013}
  {Grounding words in perception and action: computational insights}.
\newblock \emph{Trends in Cognitive Sciences}, 9(8):389 -- 396.

\bibitem[{{Samek} et~al.(2017){Samek}, {Binder}, {Montavon}, {Lapuschkin}, and
  {Müller}}]{samek2016evaluating}
W.~{Samek}, A.~{Binder}, G.~{Montavon}, S.~{Lapuschkin}, and K.~{Müller}.
  2017.
\newblock \href {https://doi.org/10.1109/TNNLS.2016.2599820} {Evaluating the
  visualization of what a deep neural network has learned}.
\newblock \emph{IEEE Transactions on Neural Networks and Learning Systems},
  28(11):2660--2673.

\bibitem[{Shi et~al.(2020)Shi, Zhao, Boyd-Graber, Daum{\'e}~III, and
  Lee}]{shi-etal-2020-potential}
Tianze Shi, Chen Zhao, Jordan Boyd-Graber, Hal Daum{\'e}~III, and Lillian Lee.
  2020.
\newblock \href {https://doi.org/10.18653/v1/2020.findings-emnlp.167} {On the
  potential of lexico-logical alignments for semantic parsing to {SQL}
  queries}.
\newblock In \emph{Findings of the Association for Computational Linguistics:
  EMNLP 2020}, pages 1849--1864, Online. Association for Computational
  Linguistics.

\bibitem[{Sorokin and Gurevych(2018)}]{sorokin-gurevych-2018-mixing}
Daniil Sorokin and Iryna Gurevych. 2018.
\newblock \href {https://doi.org/10.18653/v1/S18-2007} {Mixing context
  granularities for improved entity linking on question answering data across
  entity categories}.
\newblock In \emph{Proceedings of the Seventh Joint Conference on Lexical and
  Computational Semantics}, pages 65--75, New Orleans, Louisiana. Association
  for Computational Linguistics.

\bibitem[{Tan et~al.(2017)Tan, Wei, Ren, Lv, and Zhou}]{tan-etal-2017-entity}
Chuanqi Tan, Furu Wei, Pengjie Ren, Weifeng Lv, and Ming Zhou. 2017.
\newblock \href {https://doi.org/10.18653/v1/D17-1007} {Entity linking for
  queries by searching {W}ikipedia sentences}.
\newblock In \emph{Proceedings of the 2017 Conference on Empirical Methods in
  Natural Language Processing}, pages 68--77, Copenhagen, Denmark. Association
  for Computational Linguistics.

\bibitem[{Tenney et~al.(2019)Tenney, Xia, Chen, Wang, Poliak, McCoy, Kim,
  Durme, Bowman, Das, and Pavlick}]{tenney2018you}
Ian Tenney, Patrick Xia, Berlin Chen, Alex Wang, Adam Poliak, R.~Thomas McCoy,
  Najoung Kim, Benjamin~Van Durme, Samuel~R. Bowman, Dipanjan Das, and Ellie
  Pavlick. 2019.
\newblock \href {https://openreview.net/forum?id=SJzSgnRcKX} {What do you learn
  from context? probing for sentence structure in contextualized word
  representations}.
\newblock In \emph{7th International Conference on Learning Representations,
  {ICLR} 2019, New Orleans, LA, USA, May 6-9, 2019}. OpenReview.net.

\bibitem[{Vaswani et~al.(2017)Vaswani, Shazeer, Parmar, Uszkoreit, Jones,
  Gomez, Kaiser, and Polosukhin}]{DBLP:conf/nips/VaswaniSPUJGKP17}
Ashish Vaswani, Noam Shazeer, Niki Parmar, Jakob Uszkoreit, Llion Jones,
  Aidan~N. Gomez, Lukasz Kaiser, and Illia Polosukhin. 2017.
\newblock \href
  {https://proceedings.neurips.cc/paper/2017/hash/3f5ee243547dee91fbd053c1c4a845aa-Abstract.html}
  {Attention is all you need}.
\newblock In \emph{Advances in Neural Information Processing Systems 30: Annual
  Conference on Neural Information Processing Systems 2017, December 4-9, 2017,
  Long Beach, CA, {USA}}, pages 5998--6008.

\bibitem[{Wang et~al.(2020{\natexlab{a}})Wang, Shin, Liu, Polozov, and
  Richardson}]{wang-etal-2020-rat}
Bailin Wang, Richard Shin, Xiaodong Liu, Oleksandr Polozov, and Matthew
  Richardson. 2020{\natexlab{a}}.
\newblock \href {https://doi.org/10.18653/v1/2020.acl-main.677} {{RAT-SQL}:
  Relation-aware schema encoding and linking for text-to-{SQL} parsers}.
\newblock In \emph{Proceedings of the 58th Annual Meeting of the Association
  for Computational Linguistics}, pages 7567--7578, Online. Association for
  Computational Linguistics.

\bibitem[{Wang and Jiang(2019)}]{wang-jiang-2019-explicit}
Chao Wang and Hui Jiang. 2019.
\newblock \href {https://doi.org/10.18653/v1/P19-1219} {Explicit utilization of
  general knowledge in machine reading comprehension}.
\newblock In \emph{Proceedings of the 57th Annual Meeting of the Association
  for Computational Linguistics}, pages 2263--2272, Florence, Italy.
  Association for Computational Linguistics.

\bibitem[{Wang et~al.(2020{\natexlab{b}})Wang, Shen, Yang, Quan, and
  Wang}]{wang-etal-2020-relational}
Kai Wang, Weizhou Shen, Yunyi Yang, Xiaojun Quan, and Rui Wang.
  2020{\natexlab{b}}.
\newblock \href {https://doi.org/10.18653/v1/2020.acl-main.295} {Relational
  graph attention network for aspect-based sentiment analysis}.
\newblock In \emph{Proceedings of the 58th Annual Meeting of the Association
  for Computational Linguistics}, pages 3229--3238, Online. Association for
  Computational Linguistics.

\bibitem[{Warstadt and Bowman(2020)}]{warstadt2020neural}
Alex Warstadt and Samuel~R. Bowman. 2020.
\newblock \href {http://arxiv.org/abs/2007.06761} {Can neural networks acquire
  a structural bias from raw linguistic data?}
\newblock \emph{CoRR}, abs/2007.06761.

\bibitem[{Wolf et~al.(2020)Wolf, Debut, Sanh, Chaumond, Delangue, Moi, Cistac,
  Rault, Louf, Funtowicz, Davison, Shleifer, von Platen, Ma, Jernite, Plu, Xu,
  Le~Scao, Gugger, Drame, Lhoest, and Rush}]{wolf-etal-2020-transformers}
Thomas Wolf, Lysandre Debut, Victor Sanh, Julien Chaumond, Clement Delangue,
  Anthony Moi, Pierric Cistac, Tim Rault, Remi Louf, Morgan Funtowicz, Joe
  Davison, Sam Shleifer, Patrick von Platen, Clara Ma, Yacine Jernite, Julien
  Plu, Canwen Xu, Teven Le~Scao, Sylvain Gugger, Mariama Drame, Quentin Lhoest,
  and Alexander Rush. 2020.
\newblock \href {https://doi.org/10.18653/v1/2020.emnlp-demos.6} {Transformers:
  State-of-the-art natural language processing}.
\newblock In \emph{Proceedings of the 2020 Conference on Empirical Methods in
  Natural Language Processing: System Demonstrations}, pages 38--45, Online.
  Association for Computational Linguistics.

\bibitem[{Wu et~al.(2020)Wu, Chen, Kao, and Liu}]{wu-etal-2020-perturbed}
Zhiyong Wu, Yun Chen, Ben Kao, and Qun Liu. 2020.
\newblock \href {https://doi.org/10.18653/v1/2020.acl-main.383} {Perturbed
  masking: Parameter-free probing for analyzing and interpreting {BERT}}.
\newblock In \emph{Proceedings of the 58th Annual Meeting of the Association
  for Computational Linguistics}, pages 4166--4176, Online. Association for
  Computational Linguistics.

\bibitem[{Yu et~al.(2018{\natexlab{a}})Yu, Li, Zhang, Zhang, and
  Radev}]{yu-etal-2018-typesql}
Tao Yu, Zifan Li, Zilin Zhang, Rui Zhang, and Dragomir Radev.
  2018{\natexlab{a}}.
\newblock \href {https://doi.org/10.18653/v1/N18-2093} {{T}ype{SQL}:
  Knowledge-based type-aware neural text-to-{SQL} generation}.
\newblock In \emph{Proceedings of the 2018 Conference of the North {A}merican
  Chapter of the Association for Computational Linguistics: Human Language
  Technologies, Volume 2 (Short Papers)}, pages 588--594, New Orleans,
  Louisiana. Association for Computational Linguistics.

\bibitem[{Yu et~al.(2018{\natexlab{b}})Yu, Zhang, Yang, Yasunaga, Wang, Li, Ma,
  Li, Yao, Roman, Zhang, and Radev}]{yu-etal-2018-spider}
Tao Yu, Rui Zhang, Kai Yang, Michihiro Yasunaga, Dongxu Wang, Zifan Li, James
  Ma, Irene Li, Qingning Yao, Shanelle Roman, Zilin Zhang, and Dragomir Radev.
  2018{\natexlab{b}}.
\newblock \href {https://doi.org/10.18653/v1/D18-1425} {{S}pider: A large-scale
  human-labeled dataset for complex and cross-domain semantic parsing and
  text-to-{SQL} task}.
\newblock In \emph{Proceedings of the 2018 Conference on Empirical Methods in
  Natural Language Processing}, pages 3911--3921, Brussels, Belgium.
  Association for Computational Linguistics.

\bibitem[{Zelle and Mooney(1996)}]{zelle1996learning}
John~M. Zelle and Raymond~J. Mooney. 1996.
\newblock \href {http://www.aaai.org/Library/AAAI/1996/aaai96-156.php}
  {Learning to parse database queries using inductive logic programming}.
\newblock In \emph{Proceedings of the Thirteenth National Conference on
  Artificial Intelligence and Eighth Innovative Applications of Artificial
  Intelligence Conference, {AAAI} 96, {IAAI} 96, Portland, Oregon, USA, August
  4-8, 1996, Volume 2}, pages 1050--1055. {AAAI} Press / The {MIT} Press.

\bibitem[{Zettlemoyer and Collins(2005)}]{zettlemoyer2005learning}
Luke Zettlemoyer and Michael Collins. 2005.
\newblock \href
  {https://dslpitt.org/uai/displayArticleDetails.jsp?mmnu=1\&smnu=2\&article\_id=1209\&proceeding\_id=21}
  {Learning to map sentences to logical form: Structured classification with
  probabilistic categorial grammars}.
\newblock In \emph{{UAI} '05, Proceedings of the 21st Conference in Uncertainty
  in Artificial Intelligence, Edinburgh, Scotland, July 26-29, 2005}, pages
  658--666. {AUAI} Press.

\bibitem[{Zhou et~al.(2019)Zhou, Kalantidis, Chen, Corso, and
  Rohrbach}]{DBLP:conf/cvpr/ZhouKCCR19}
Luowei Zhou, Yannis Kalantidis, Xinlei Chen, Jason~J. Corso, and Marcus
  Rohrbach. 2019.
\newblock \href {https://doi.org/10.1109/CVPR.2019.00674} {Grounded video
  description}.
\newblock In \emph{{IEEE} Conference on Computer Vision and Pattern
  Recognition, {CVPR} 2019, Long Beach, CA, USA, June 16-20, 2019}, pages
  6578--6587. Computer Vision Foundation / {IEEE}.

\bibitem[{Zhu et~al.(2016)Zhu, Groth, Bernstein, and
  Fei{-}Fei}]{DBLP:conf/cvpr/ZhuGBF16}
Yuke Zhu, Oliver Groth, Michael~S. Bernstein, and Li~Fei{-}Fei. 2016.
\newblock \href {https://doi.org/10.1109/CVPR.2016.540} {Visual7w: Grounded
  question answering in images}.
\newblock In \emph{2016 {IEEE} Conference on Computer Vision and Pattern
  Recognition, {CVPR} 2016, Las Vegas, NV, USA, June 27-30, 2016}, pages
  4995--5004. {IEEE} Computer Society.

\end{thebibliography}
